\UseRawInputEncoding
\documentclass[]{IEEEtran}
\pdfoutput=1%For ARXIV
\usepackage{graphicx}
\usepackage{amsmath}
\usepackage{amssymb}		
\usepackage{amsmath,amsfonts}
\usepackage{algorithmic}
\usepackage{array}
\usepackage[caption=false,font=normalsize,labelfont=sf,textfont=sf]{subfig}
\usepackage{textcomp}
\usepackage{stfloats}
\usepackage{url}
\usepackage{verbatim}
\usepackage{graphicx}
\usepackage{cite}

\usepackage{times}
\usepackage{soul}
\usepackage{url}
\usepackage{graphicx}
\usepackage{amsmath}
\usepackage{amsthm}
\usepackage{booktabs}
\usepackage{algorithm}
\usepackage{algorithmic}
\usepackage[switch]{lineno}
\usepackage{makecell}
\usepackage{multirow}
\usepackage{makecell}
\usepackage{amssymb}
\usepackage{pifont} 
\usepackage{float}
\usepackage{MnSymbol}
\usepackage{xcolor}

\usepackage{colortbl}  %彩色表格需要加载的宏包
\usepackage{xcolor}
\usepackage{array}
\begin{document}

\title{Caformer: Rethinking Time Series Analysis from Causal Perspective}
\author{Kexuan Zhang, Xiaobei Zou, Yang Tang
        % <-this % stops a space
% \thanks{This work was supported by the National Natural Science Foundation of China (62293502, 62293504, 62173147). (Corresponding author: Yang Tang.)}
\thanks{Kexuan Zhang, Xiaobei Zou and Yang Tang are with the Key Laboratory of
Smart Manufacturing in Energy Chemical Process, Ministry of Education, East
China University of Science and Technology, Shanghai 200237, China (e-mail:
kexuanzhang123@gmail.com; xbeizou@gmail.com; yangtang@ecust.edu.cn).}
\thanks{This work was supported by National Natural Science Foundation of China (62233005, 62293502), the Programme of Introducing Talents of Discipline to Universities (the 111 Project) under Grant B17017, Fundamental Research Funds for the Central Universities (222202317006) and Shanghai AI Lab.}}


\maketitle

\begin{abstract}
Time series analysis is a vital task with broad applications in various domains. However, effectively capturing cross-dimension and cross-time dependencies in non-stationary time series poses significant challenges, particularly in the context of environmental factors. The spurious correlation induced by the environment confounds the causal relationships between cross-dimension and cross-time dependencies. In this paper, we introduce a novel framework called Caformer (\underline{\textbf{Ca}}usal Trans\underline{\textbf{former}}) for time series analysis from a causal perspective. Specifically, our framework comprises three components: Dynamic Learner, Environment Learner, and Dependency Learner. The Dynamic Learner unveils dynamic interactions among dimensions, the Environment Learner mitigates spurious correlations caused by environment with a back-door adjustment, and the Dependency Learner aims to infer robust interactions across both time and dimensions. Our Caformer demonstrates consistent state-of-the-art performance across five mainstream time series analysis tasks, including long- and short-term forecasting, imputation, classification, and anomaly detection, with proper interpretability.
\end{abstract}


\section{Introduction}
\IEEEPARstart{T}ime series analysis holds immense practical value in real-world fields, including finance \cite{iTansformer,8936546}, climate science \cite{9669023}, and traffic management \cite{9829541}. It has attained growing attention among researchers for its broad practical value \cite{TimesNet}. However, with complex temporal dependencies and dimension interactions involved, the analysis of time series is challenging.\\
Numerous efforts \cite{iTansformer,AnomalyTransformer} have been devoted to struggling with the modeling of dependencies. Within this spectrum, Transformer \cite{vaswani2017attention} and its variants have shown notable capabilities in capturing cross-time dependency \cite{wen2022transformers}. Furthermore, there is a growing emphasis on interactions between cross-dimension and cross-time dependencies for a more comprehensive understanding \cite{crossformer}. 
\begin{figure}[ht]
    \centering
    \includegraphics[height=6cm]{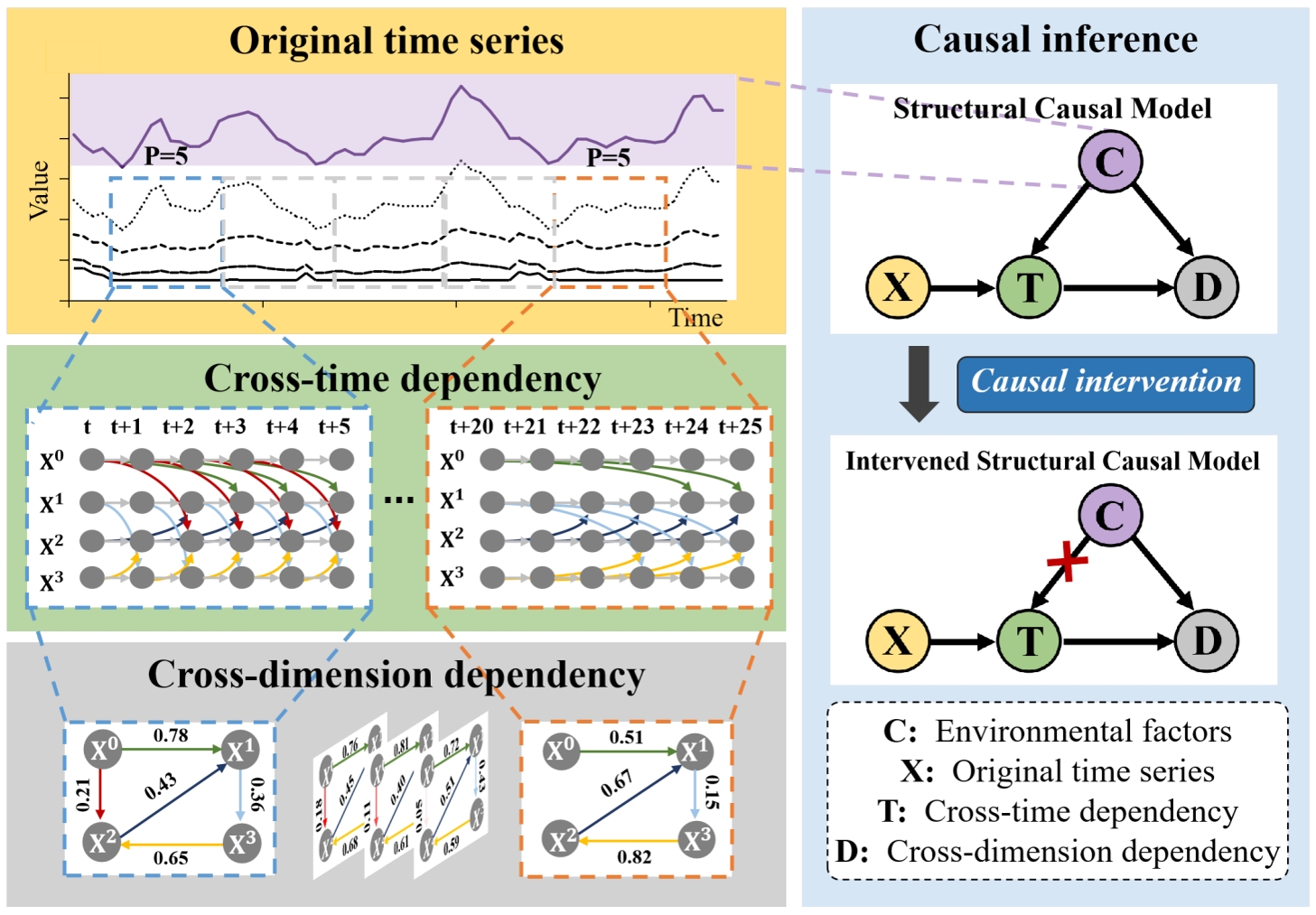}
    \caption{Dynamic interactions within time series from a causal perspective. The original time series $\textbf{X}$ comprises four variables, represented by various types of black lines, along with a purple line symbolizing environmental factors. The cross-time dependency is extracted from two time periods with the same patch size $P=5$, while the cross-dimension dependency is inferred from the corresponding cross-time dependency. The structural causal model shows the causality between environmental factors, original time series, cross-dimension and cross-time dependencies.\\}
    \label{motivation}
    \vspace{-0.5cm}
\end{figure}\\
\indent However, such dependencies are dynamic and susceptible to external effects. As the inherent property of time series, non-stationary leads to continual changes in statistical characteristics and joint distributions, which can be manifested as dynamic variations in dependencies \cite{liu2022non}. Fig. \ref{motivation} demonstrates our understanding of the dynamic interactions within time series. The cross-dimension and cross-time dependencies can be derived successively from the time series. Cross-dimension dependency captures interactions among dimensions, while cross-time dependency represents the dynamic propagation and interactions among dimensions over time. \\
\indent It is worth noting that despite the significant strides in handling complex interactions in previous research \cite{Autoformer,iTansformer}, a comprehensive exploration into the causal chains underlying these dynamic dependencies remains incomplete. With regard to cross-time dependency, prevailing approaches \cite{Autoformer,PatchTST} strive to embed data points from all dimensions at the same time step into a feature vector \cite{crossformer}, aiming to uncover inner dependencies among different time steps. Nevertheless, it is crucial not to overlook the impact of external factors that propagate over time, as they may introduce biases. Concerning cross-dimension dependency, prior methods \cite{crossformer,iTansformer} grapple with explicit capturing of dependencies from each individual dimension-specific embedding. However, the learned dependency is approached from a correlation perspective, ignoring the underlying causal graph, where the interaction strength within the graph can indeed vary.\\
\indent To address the aforementioned challenges, we propose a general model termed Caformer (\underline{\textbf{Ca}}usal Trans\underline{\textbf{former}}) for time series analysis with a task-independent backbone and different heads for specific downstream tasks. The core idea is to obtain a deep understanding of the dynamics that characterize causal relationships between cross-dimension dependency and cross-time dependency, especially in the presence of environmental factors. Specifically, the main contributions are as follows:
\begin{itemize}
    \item We establish a structural causal model to explain the underlying mechanisms of interactions in time series and demonstrate the problem of dependency varying from a causal perspective. A novel framework called Caformer is proposed for more accurate and interpretable time-series analysis.
    \item The Dynamic Learner is proposed to uncover the dynamic underlying causal relations concerning cross-dimension dependency. The Environment Learner quantifies and stratifies the environmental factors with the back-door adjustment for a robust cross-time dependency under varying environments. The Dependency Learner is designed to infer the robust interactions between these two dependencies.
    \item Caformer achieves consistent state-of-the-art performance on multiple mainstream time series analysis tasks, demonstrating its excellent generalization ability for different tasks and its proper interpretability.
\end{itemize}
\section{Related work}
\subsection{Time series analysis.}
In recent years, there has been extensive research into time series analysis, with a particular focus on investigating cross-dimension and cross-time dependencies. Table \ref{Transformer list} summarizes the distinctive characteristics of the existing time series methods. The RNN-based methods \cite{9127499} capture cross-time dependency with the recurrent structure. The MLP-based methods \cite{ekambaram2023tsmixer,DLinear} encode the cross-time dependency with parameterized MLP layers. The TCN-based methods \cite{zhang2023trid} obtain the cross-time dependency with the convolution kernels. Meanwhile, the Transformer-based methods \cite{Autoformer,FEDformer} embed data points from the same time step and learn the cross-time dependency with the well-designed attention mechanisms. Notably, there has been a recent surge of interest \cite{crossformer,iTansformer} in cross-dimension dependency with an enhanced understanding of time series. Consequently, developing a comprehensive understanding of dependencies within time series data is crucial for effective time series analysis. In this paper, we address both cross-dimension and cross-time dependencies and refine the learning of such dependencies from a dynamic perspective.

\begin{table}[ht]
    \scriptsize
    \centering
    \renewcommand\arraystretch{1.15}
    \caption{Different characteristics of the popular Transformer-based models. CDD (\underline{\textbf{C}}ross-\underline{\textbf{D}}imension \underline{\textbf{D}}ependency) and CTD (\underline{\textbf{C}}ross-\underline{\textbf{T}}ime \underline{\textbf{D}}ependency) indicate whether the methods learn cross-dimension and cross-time dependencies, respectively. Dynamic shows the condition of the learning of dynamic interactions.\label{Transformer list} }
    \begin{tabular}{c|c|c|c|c}
    \toprule
    Model name&Task&CDD&CTD&Dynamic\\
    \midrule
    \cite{9158560}&Imputation&\ding{55}&\ding{51}&\ding{55}\\
    \cite{GTN}&Classification&\ding{55}&\ding{51}&\ding{55}  \\
    \cite{SVP-T}&Classification&\ding{51}&\ding{51}&\ding{55}  \\
    \cite{10316684}&Anomaly Detection&\ding{51}&\ding{51}&\ding{55}\\
    \cite{AnomalyTransformer}&Anomaly Detection&\ding{55}&\ding{51}&\ding{55}\\
    \cite{informer}&Forecasting&\ding{55}&\ding{51}&\ding{55}\\
    \cite{Autoformer}&Forecasting&\ding{55}&\ding{51}&\ding{55}\\
    \cite{ekambaram2023tsmixer}&Forecasting&\ding{55}&\ding{51}&\ding{55}\\
    \cite{crossformer}&Forecasting&\ding{51}&\ding{51}&\ding{55}  \\
    \cite{PatchTST}&Forecasting&\ding{55}&\ding{51}&\ding{55}\\
    \cite{iTansformer}&Forecasting&\ding{51}&\ding{55}&\ding{55}\\
    \bottomrule
    \end{tabular}
\end{table}
\begin{figure*}[ht]
    \centering
    \includegraphics[height=8.5cm]{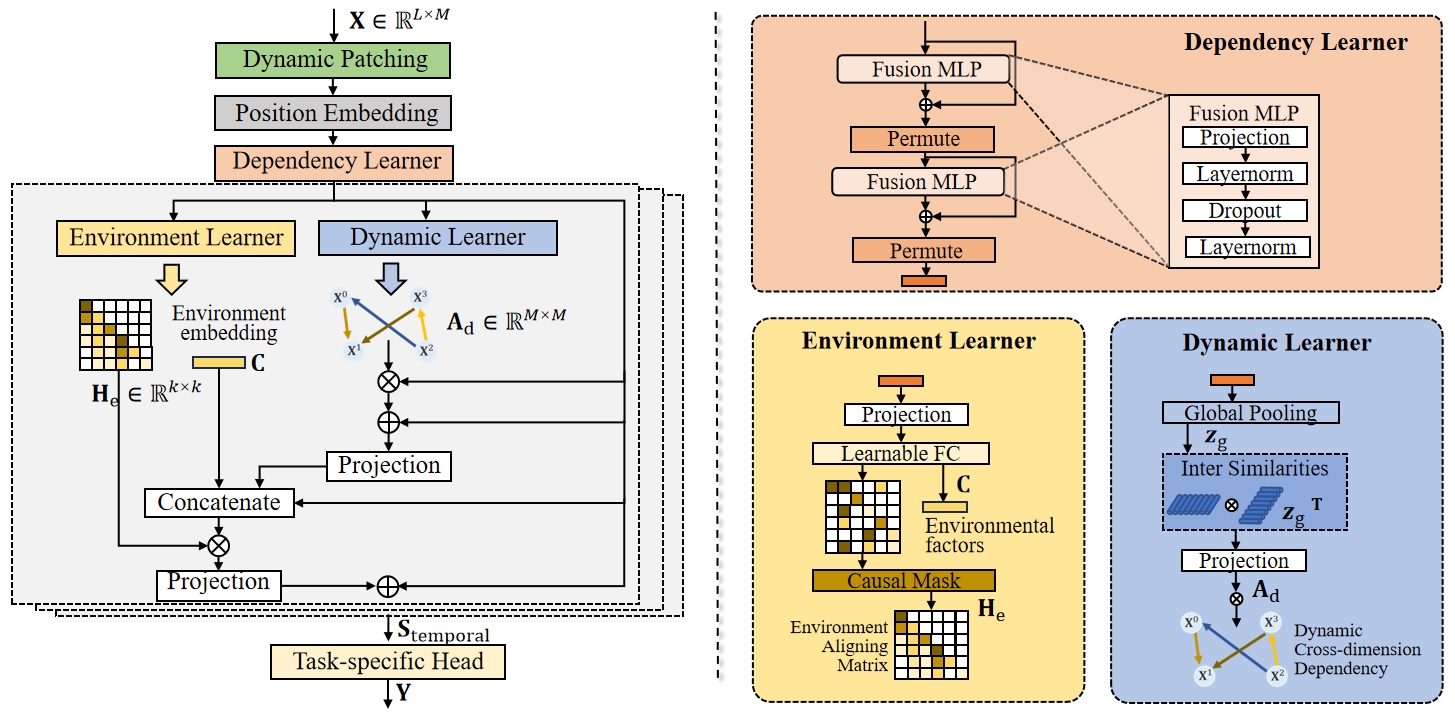}
    \caption{The architecture of Caformer (\underline{\textbf{Ca}}usal Trans\underline{\textbf{former}}). It comprises three components: Dependency Learner, Environment Learner, and Dynamic Learner.}
    \label{structure}
\end{figure*}
\subsection{Dynamics in time series.} The identification of deterministic dynamics within observed time series provides valuable insights into underlying physical processes, especially on shorter timescales \cite{barahona1996detection}. Interactions among components are widespread in natural systems, contributing to the complexity of time series analysis. While significant efforts have been devoted to learning dynamical models of interactions using techniques like graph neural networks \cite{zevcevic2021relating,lowe2022amortized} and attention mechanisms \cite{goyal2021recurrent}, it's noteworthy that dynamic relationships are often overlooked in time series analysis. Existing methods primarily focus on learning dependencies across multiple dimensions, neglecting explicit modeling of dynamic interactions. In this paper, we propose an approach that explicitly learns the dynamics within the time series to enhance the quality of learned dependencies.
\subsection{Causal inference.} Causal inference plays a crucial role in uncovering the causal structure of systems and quantifying causal effects. It achieves this by integrating domain knowledge, machine models, and observational or interventional data \cite{runge2023causal}. Recognizing that "Correlation is not causation," recent advancements in deep learning models have incorporated causal inference to infer causal effects in various domains, including computer vision \cite{zhang2020causal}, natural language processing \cite{CausalInNlp}, and robotics \cite{castri2023continual}. Spurious correlation refers to situations where a misleading correlation between two variables arises due to the influence of a third causal variable \cite{haig2003spurious}. These accidental correlations often result from factors such as sample selection bias, making it challenging to discern true causation. With the help of causal inference, such spurious bias can be eliminated, and the desired model effects can be disentangled. In our work, we leverage the structural causal model developed by Pearl \cite{pearl2000models} to analyze the relationships among the original time series, the cross-dimension dependency, the cross-time dependency, and the environmental factors.
\section{Caformer}
To comprehensively unveil cross-dimension and cross-time dependencies within time series, a novel framework called Caformer is proposed, depicted in Fig. \ref{structure}. \\
\indent Given $\textbf{X} \in \mathbb{R}^{M\times L}$, representing the multivariate time series with a length of $L$ and $M$ recorded dimensions, each $\textbf{X}_{t}^{i}$ can be expressed as follows: 
\begin{equation}
    \textbf{X}_{t}^{i}:=\text{f}_{i}(\text{pa}(\textbf{X}_{t}^{i}),\boldsymbol{\eta}_{t}^{i}),
    \label{pa}
\end{equation}
where $\text{f}_{i}(\cdot)$ is the nonlinear functional causal mechanism determining the value of $\textbf{X}_{t}^{i}$ based on its causal parents $\text{pa}(\textbf{X}_{t}^{i})$ and the independent noise $\boldsymbol{\eta}_{t}^{i}$. The causal parents $\text{pa}(\textbf{X}_{t}^{i})$ form a subset of $\{\textbf{X}_{t},\dots, \textbf{X}_{t-\tau_{\text{max}}}\}\backslash\{\textbf{X}_{t}^{i}\}$ with time lag $\tau_{\text{max}}\geq0$ \cite{runge2023causal}, indicating all the causes of $\textbf{X}_{t}^{i}$ except for itself. As direct causes, $\text{pa}(\textbf{X}_{t}^{i})$ propagates their effects on $\textbf{X}_{t}^{i}$ with finite time lags of at most $\tau_{\text{max}}$ time steps. The noise $\boldsymbol{\eta}_{t}^{i}$ encompasses the effects of all the external factors. Combining this insight with Fig. \ref{motivation}, it can be deduced that $\text{f}_{i}(\cdot)$ reflects the causal interactions among dimensions as the cross-dimension dependency, while $\text{pa}(\cdot)$ captures the propagation among dimensions from the temporal perspective, embodying cross-time dependency. Additionally, the influence of external factors $\boldsymbol{\eta}_{t}^{i}$ cannot be neglected. Their impact on time series can contribute to inevitable distribution variations and dynamic changes. Therefore, it is of great importance to have a comprehensive understanding of cross-dimension and cross-time dependencies to reveal the underlying causal mechanisms within time series.
% \begin{figure*}[htbp]
%     \centering
%     \includegraphics[height=8.5cm]{structure2.png}
%     \caption{The architecture of Caformer.}
%     \label{structure}
% \end{figure*}
\subsection{Structural Causal Model}
We construct a structural causal model \cite{pearl2000models} to elucidate the causal relationships among four variables: the original time series $\textbf{X}$, the environmental factors $\textbf{C}$, the cross-time dependency $\textbf{T}$, and the cross-dimension dependency $\textbf{D}$, in Fig. \ref{motivation}. The directed edges in the graph denote the causal relationships between nodes. The specifics of these cause-effect relationships are elaborated below:\\
\indent $\textbf{X}\xrightarrow{}\textbf{T}\xrightarrow{}\textbf{D}$. The cross-time dependency can be discerned from time series, whereas the cross-dimension dependency represents the dynamic interactions concealed within the cross-time dependency. These paths elucidate the inferential process of distinct dependencies within time series.\\
\indent $\textbf{T}\xleftarrow{}\textbf{C}\xrightarrow{}\textbf{D}$. Both the cross-time dependency $\textbf{T}$ and the cross-dimension dependency $\textbf{D}$ are susceptible to the influence of environmental factors. In the context of cross-time dependency, the environment introduces dynamic dependencies through non-stationary distributions. These distributions induce shifts in time lags at the time level, causing a delay or advancement in the point of action. Concerning cross-dimension dependency, the environment's impact on one dimension alters the direction and strength of interactions among all dimensions, leading to a fluctuating cross-dimension dependency.\\
\indent However, the back-door path \indent $\textbf{T}\xleftarrow{}\textbf{C}\xrightarrow{}\textbf{D}$ between $\textbf{T}$ and $\textbf{D}$ can bring the spurious correlation when inferring $\textbf{D}$ from $\textbf{T}$. Changes in cross-time dependency may be erroneously interpreted as a shift in the strength of the action, introducing additional non-causal cross-dimension dependency due to variations in propagation through different time lags.
\subsection{Causal Intervention via Back-door Adjustment}
As stated above, the inference of the cross-dimension and cross-time dependencies can be confounded by the inevitable environmental factors. Fortunately, with the guidance of the back-door criterion \cite{pearl2000models}, the edge $\textbf{C}\xrightarrow{}\textbf{T}$ can be cut off to eliminate the confounding between $\textbf{T}$ and $\textbf{D}$. Given the stratified environmental factors, the back-door adjustment can be conducted as in Eq. (\ref{backdoor}).
\begin{equation}
    \mathcal{P}(\textbf{D} |\textit{do}(\textbf{T})) = \sum_{i}^{n}\mathcal{P}(\textbf{D}|\textbf{T},\textbf{c}_{i})\mathcal{P}(\textbf{c}_{i}),
    \label{backdoor}
\end{equation}
where $n$ indicates the number of the stratified environmental factors $\left\{\textbf{c}_{i}\right\}_{i=1}^{n}$, and $\textit{do}(\textbf{T})$ denotes the do-calculus on $\textbf{T}$, cutting off all the edges pointing to $\textbf{T}$. It can be inferred from Eq. (\ref{backdoor}) that the cross-dimension dependency is calculated by incorporating each environmental factor $\textbf{c}_{i}$ with the cross-time dependency and applying weights through $\mathcal{P}(\textbf{c}_{i})$. \\
\indent Consequently, $\text{pa}(\cdot)$ in Eq. (\ref{pa}) is equivalent to the inference of $\mathcal{P}(\textbf{T})$, while $\mathcal{P}(\textbf{D})$ can be modified as $\mathcal{P}(\textbf{D}|\textit{do}(\textbf{T}))$ concerning $\text{f}_{i}(\cdot)$ to eliminate spurious correlations induced by the environment. In this paper, the Dynamic Learner is proposed to reveal the underlying dynamic interactions among dimensions; the Environment Learner is designed to quantify and stratify the environmental factors with back-door adjustments; and the Dependency Learner is introduced to grasp robust interactions among these two dependencies.
\vspace{-1pt}
\subsection{Dependency Learner and Dynamic Learner}
\begin{figure}[ht]
    \centering
    \includegraphics[height=3.8cm]{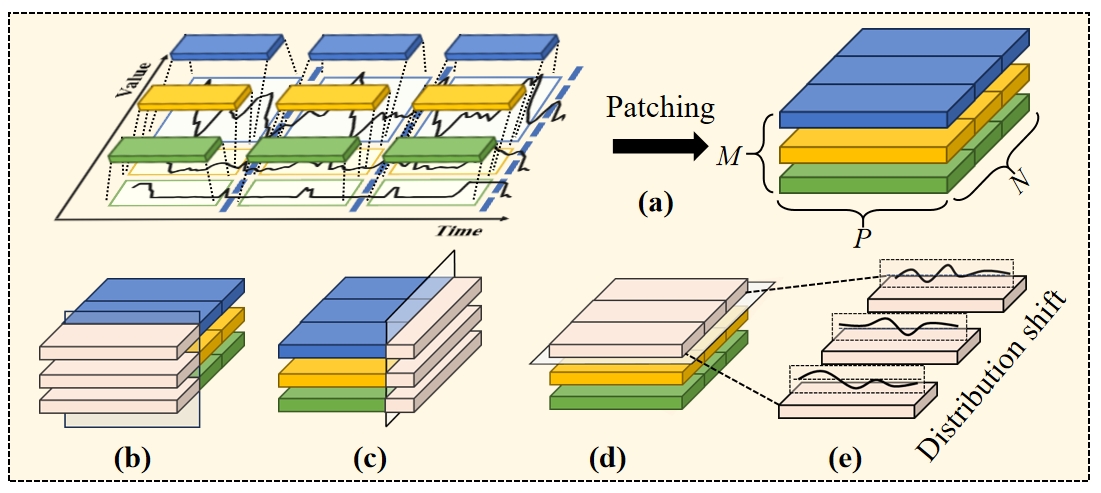}
    \caption{The illustration of patching process and the dependencies among patched time series.}
    \label{intro}
\end{figure}

Since the underlying causal relationships within time series are not constant over time due to non-stationarity, a fixed structure cannot conclusively represent the entire time series. However, individual time point in time series may be easily inferred from their neighbors and might not contain significant semantic information. Hence, we make the assumption that the causal relationships within a specific period of time remain stable and try to learn the dimension-independent embedding with the patches of series. The illustration of the patching process is shown in Fig. \ref{intro} (a).
\\
\textbf{Dependency Learner.} With the patched data $\textbf{X}_{\text{p}} \in \mathbb{R}^{M \times P\times N}$, where $N=\frac{L-P}{S}+2$ is the number of patches, and both $P$ and $S$ are hyper-parameters concerning patching, the obtained embedding can represent $N$ stacked time patches with different underlying causal relationships. The Dependency Learner is proposed to model the interactions between cross-dimension and cross-time dependencies. As depicted in Fig. \ref{intro}, these embeddings encompass various dependencies. Fig. \ref{intro} (b) indicates the cross-dimension dependency over a period of time; Fig. \ref{intro} (c) shows the cross-dimension dependency at the same time step; Fig. \ref{intro} (d) refers to the cross-time dependency among patches. Since the points of the same time step may basically represent completely different physical meanings recorded by inconsistent measurements \cite{iTansformer}, only the dependencies in Fig. \ref{intro} (b) and (d) are learned with Dependency Learner in Fig. \ref{structure}. The in-patch normalization is also proposed to eliminate the distribution shift between patches as shown in Fig. \ref{structure} (e). Finally, the interactions between cross-dimension and cross-time dependencies $\textbf{I}_{\text{de}} \in \mathbb{R}^{N \times M\times E}$ are obtained, where $E$ denotes the size of the embeddings.\\
\textbf{Dynamic Learner.} The Dynamic Learner is proposed to infer the underlying dynamics in different patch series for cross-dimension dependency. The global dimension information $\textbf{z}_{\text{g}} \in \mathbb{R}^{N \times M\times 1}$ is first obtained from the learned dependencies $\textbf{I}_{\text{de}}$ in Dependency Learner. The inter-similarity $\textbf{A}_{\text{d}} \in \mathbb{R}^{N \times M\times M}$ is then calculated for the expression of the dynamic interactions between dimensions.
\begin{align}
    &\textbf{z}_{\text{g}}=\text{FC}(\text{GPooling}(\textbf{I}_{\text{de}})),\\
    &\textbf{A}_{\text{d}} =\text{Norm}(\phi(\frac{\textbf{z}_{\text{g}}\textbf{z}_{\text{g}}^\mathsf{T}}{\sqrt{\alpha}})),
\end{align}
where GPooling$(\cdot)$ denotes the pooling function, FC $(\cdot)$ is the Fully Connected layer, $\phi(\cdot)$ is the Softmax function, Norm $(\cdot)$ is the normalization function, and $\alpha$ is the given parameter for normalization. With $\textbf{A}_{\text{d}}$, the cross-dimension dependency $\textbf{D} \in \mathbb{R}^{N \times M\times M}$ can be strengthened as follows:
\begin{equation}
    \textbf{D} = \textbf{A}_{\text{d}}\textbf{I}_{\text{de}}.
\end{equation}
\subsection{Environment Learner}
As stated above, explicit modeling of environmental factors can promote the elimination of spurious correlations between cross-dimension and cross-time dependencies. The Environment Learner is proposed to quantify and stratify the environmental factors for both $\mathcal{P}(\textbf{c}_{i})$ and $\left\{\textbf{c}_{i}\right\}_{i=1}^{n}$, as shown in Fig. \ref{structure}. With the dependencies $\textbf{I}_{\text{de}}$ learned from Dependency Learner, the environmental factors $\textbf{C}$ can be inferred as follows:
\begin{align}
    &\textbf{S}_{\text{e}}= \text{F}_{\text{e}}(\textbf{I}_\text{de}),\\
    &\textbf{C}=\boldsymbol{\alpha}_{2}(\text{ReLU}(\boldsymbol{\alpha}_{1}(\textbf{S}_{\text{e}})+\boldsymbol{\gamma}_{1}))+\boldsymbol{\gamma}_{2},
\end{align}
where $\text{F}_{\text{e}}(\cdot$) denotes the modeling of the latent representation of environmental factors, $\boldsymbol{\alpha}_{1}$, $\boldsymbol{\alpha}_{2}$ and $\boldsymbol{\gamma}_{1}$, $\boldsymbol{\gamma}_{2}$ are learnable model parameters. ReLU($\cdot$) indicates the ReLU function. After modeling $\left\{\textbf{c}_{i}\right\}_{i=1}^{n}$, $\textbf{H}_{e}\in \mathbb{R}^{k\times k}$ is used to denote $\mathcal{P}(\textbf{c}_{i})$, where $k$ is the hyper-parameter determining the stratifying condition.
\begin{equation}
    \textbf{H}_{\text{e}}= \text{Norm}(\phi(\frac{\text{Proj}(\textbf{C})\text{Proj}(\textbf{C})^\mathsf{T}}{\sqrt{\beta}})),
\end{equation}
where Proj($\cdot$) maps the the environmental factors to $k$ dimensions, and $\beta$ is the predefined parameter for normalization. Instead of directly using $\textbf{H}_{\text{e}}$, a causal mask function $\text{M}_{\text{c}}(\cdot)$ is used to insure the potential embedding consistent with causality, since future value can't have an impact on history information. $\mathcal{P}(\textbf{c}_{i})$ can ultimately be expressed as $\textbf{H}_{\text{ce}}=\text{M}_{c}(\textbf{H}_{\text{e}})$.\\
\indent The environmental factors, the cross-time dependency, and the intervened cross-dimension dependency are concentrated together, strengthening the learned temporal features $\textbf{S}_{\text{temporal}}$ with fully learned dependencies.
\begin{align}
    &\textbf{T}= \text{Norm}(\textbf{H}_{\text{ce}}\text{FC}(\text{Concat}(\textbf{D}, \textbf{C},\textbf{I}_\text{de}))),\\
    &\textbf{S}_{\text{temporal}}=\textbf{T}+\textbf{I}_\text{de},
\end{align}
\indent Finally, $\textbf{S}_{\text{temporal}}$ can be imported into any task-specific head, corresponding to different analysis tasks for the specific label $\textbf{Y}$.
\begin{table*}[htbp]
    \scriptsize
    \centering
    \renewcommand\arraystretch{1.15}
    \caption{Summary of experiment benchmarks and corresponding metrics.}
    \begin{tabular}{c|c|c}
        \toprule
         Tasks&Benchmarks&Metrics  \\
         \midrule
         Long-term forecasting&ETT (4 subsets), Electricity, Traffic, Weather, Exchange, ILI&MSE, MAE\\
         Short-term forecasting&M4 (6 subsets)&SMAPE, MASE, OWA\\
         Imputation&ETT (4 subsets), Electricity, Weather&MSE, MAE\\
         Classification&UEA (10 subsets)&Classification Accuracy\\
         Anomaly detection&SMD, MSL, SMAP, SWaT, PSM&Precision, Recall, F1-score\\
         \bottomrule
    \end{tabular}
    \label{Summary of experiment benchmarks}
\end{table*}
\section{Experiments}
To validate the effectiveness of Caformer, comprehensive experiments are conducted on five popular time series analysis tasks, including long- and short-term forecasting, imputation, classification, and anomaly detection, utilizing benchmark datasets. To ensure a fair comparison, we adhere to the well-established experimental setups and implementations of task-general baselines in \cite{TimesNet}. \\
\indent \textbf{Implementation} Table \ref{Summary of experiment benchmarks} summarizes the benchmarks. All the experiments are implemented in PyTorch and conducted on a single NVIDIA GeForce RTX 3090 GPU with 24GB memory. \\
To ensure the generalization of the model, the number of layers, the size of the matrix, and the patch size are designed respectively, considering the different characteristics of the models and datasets. In order to provide a fair comparison, we exclusively assess the base models' capabilities while maintaining the identical input embedding and final projection layer across them. The modular architecture of learning a task-independent `backbone' to capture the temporal features and different `heads' for specific downstream tasks is adopted. \\
\indent We compare our model with the latest models in the time series community. The baselines include Transformer-based models: iTransformer \cite{iTansformer}, PatchTST \cite{PatchTST}, Crossformer \cite{crossformer}, FEDformer 
 \cite{FEDformer}, and ETSformer \cite{etsformer}; Convolution-based models: TimesNet \cite{TimesNet}; MLP-based models: DLinear \cite{DLinear}, and LightTS \cite{LightTS}.\\
\indent Concerning the metrics, both the mean square error (MSE) and mean absolute error (MAE) are used as metrics in long-term forecasting tasks. For short-term forecasting, the symmetric mean absolute percentage error (SMAPE), mean absolute scaled error (MASE) and overall weighted average (OWA) are adopted as the metrics. In imputation tasks, both the mean square error (MSE) and mean absolute error (MAE) are introduced. For classification, we use the accuracy as metrics. Concerning the anomaly detection tasks, we adopt the F1-score, which is the harmonic mean of precision and recall. All the metrics are calculated as follows:
\begin{equation}
    \text{MSE} = \frac{1}{T}\sum_{i=1}^{T}(\textbf{X}_{i}-\hat{\textbf{X}}_{i})^{2},
\end{equation}
\begin{equation}
     \text{MAE} = \frac{1}{T}\sum_{i=1}^{T}\left\vert \textbf{X}_{i}-\hat{\textbf{X}}_{i} \right\vert,
\end{equation}
\begin{equation}
    \text{SMAPE} = \frac{200}{T}\sum_{i=1}^{T}\frac{\left\vert \textbf{X}_{i}-\hat{\textbf{X}}_{i} \right\vert}{\left\vert {\textbf{X}_{i}}\right\vert + \left\vert {\hat{\textbf{X}}_{i}}\right\vert},
\end{equation}
\begin{equation}
    \text{MAPE} = \frac{100}{T}\sum_{i=1}^{T}\frac{\left\vert \textbf{X}_{i}-\hat{\textbf{X}}_{i} \right\vert}{\left\vert {\textbf{X}_{i}}\right\vert},
\end{equation}
\begin{equation}
    \text{MASE} = \frac{1}{T}\sum_{i=1}^{T}\frac{\left\vert \textbf{X}_{i}-\hat{\textbf{X}}_{i} \right\vert}{\frac{1}{T-m}\sum_{j=m+1}^{T}{\left\vert \textbf{X}_{j}-\textbf{X}_{j-m}\right\vert}},
\end{equation}
\begin{equation}
    \text{OWA} = \frac{1}{2}\left[\frac{\text{SMAPE}}{\text{SMAPE}_{\text{Naive2}}}+\frac{\text{MASE}}{\text{MASE}_{\text{Naive2}}}\right],
\end{equation}
\begin{equation}
    \text{F1-score} = \frac{2\times \text{Precision} \times \text{Recall}}{\text{Precision}+\text{Recall}},
\end{equation}
where $\textbf{X}_{i}$ and $\hat{\textbf{X}}_{i}$ are the ground truth and forecasting of the $i$-th time step in $T$ future time steps, and $m$ is the periodicity of the data. $\text{SMAPE}_{\text{Naive2}}$ and $\text{MASE}_{\text{Naive2}}$ are SMAPE and MASE of the baseline method provided in the M4 competition in \cite{M4}.
\subsection{Main Results}
\begin{figure}[ht]
    \centering
    % \hspace{-1.2cm}
    \includegraphics[height=5cm]{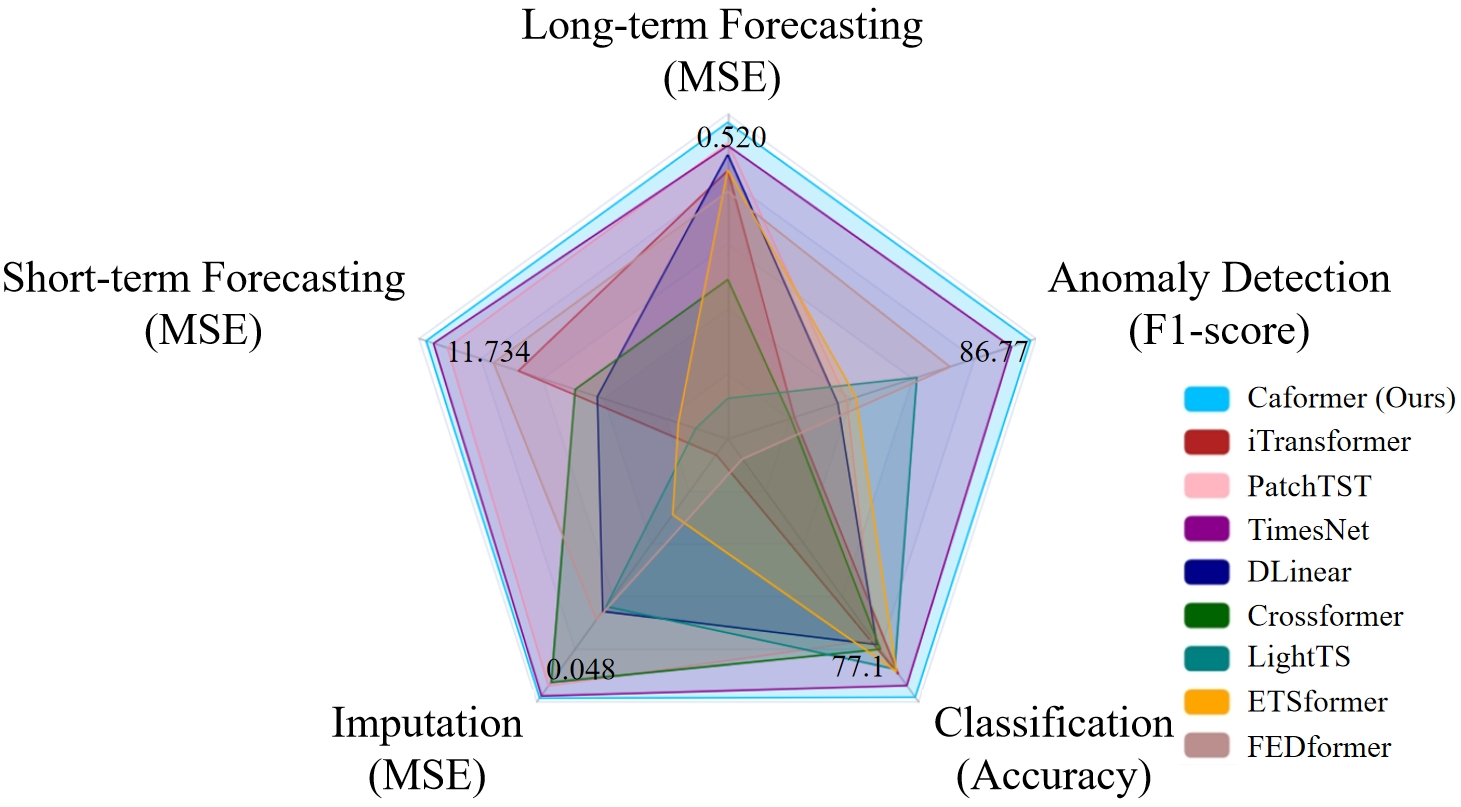}
    \caption{Comparison between models on five mainstream time series analysis tasks. A larger area indicates a better generalization result of the method across tasks.}
    \label{performance}
\end{figure}

As shown in Fig. \ref{performance}, our Caformer achieves consistent state-of-the-art performance on five mainstream analysis tasks. The details and summarized results of each task are shown in the following subsections.
\subsection{Long-term forecasting}
\textbf{Setups} Long-term forecasting is a category of time series forecasting tasks characterized by an extensive forecasting horizon as output, demanding models with robust capabilities in long-term modeling. We conduct long-term forecasting experiments on 9 popular benchmarks, including ETT \cite{informer}, Electricity\footnote{\url{https://archive.ics.uci.edu/ml/datasets/ElectricityLoadDiagrams20112014}\label{elefoot}}, Weather\footnote{\url{https://ncei.noaa.gov/data/local-climatological-data}\label{weafoot}}, ILI\footnote{\url{https://gis.cdc.gov/grasp/fluview/fluportaldashboard. html}}, Traffic\footnote{\url{http://pems.dot.ca.gov/}}, and Exchange \cite{Exchange}. The MSE and MAE are used as the metrics, and the input lengths are fixed for a fair comparison.\\
\indent \textbf{Results} As shown in Table \ref{long}, Caformer shows great performance in long-term forecasting. With a thorough grasp of interactions within time series, Caformer proficiently learns the underlying causal relationships, effectively capturing features that contribute to predicting future trends.
\begin{table*}[htbp]
    % \fontsize{7.2}{8.8}\selectfont
    \scriptsize
    \renewcommand\arraystretch{1.15}
    \setlength{\tabcolsep}{5pt}
    \centering
    \caption{Long-term forecasting task. The lengths of the input sequences are set as 36 for ILI and 96 for the others. All the results are the average of four prediction lengths, that is \{24, 36, 48, 60\} for ILI and \{96, 192, 336, 720\} for the others. A lower MSE or MAE indicates a better performance. The best results are in \textbf{bold}  and the second best are \underline{underlined}.}
    \begin{tabular}{c|c|cc|cc|cc|cc|cc|cc|cc|cc|cc}
    \toprule
    \multicolumn{2}{c|}{Models}&\multicolumn{2}{c|}{Ours}&\multicolumn{2}{c|}{\makecell[c]{iTransformer\\(2024)}}&\multicolumn{2}{c|}{\makecell[c]{PatchTST\\(2023)}}&\multicolumn{2}{c|}{\makecell[c]{TimesNet\\(2023)}}&\multicolumn{2}{c|}{\makecell[c]{Dlinear\\(2023)}}&\multicolumn{2}{c|}{\makecell[c]{Crossformer\\(2023)}}&\multicolumn{2}{c|}{\makecell[c]{LightTS\\(2022)}}&\multicolumn{2}{c|}{\makecell[c]{ETSformer\\(2022)}}&\multicolumn{2}{c}{\makecell[c]{FEDformer\\(2022)}}\\
    \midrule
    \multicolumn{2}{c|}{Metric}&\multicolumn{1}{r}{MSE}&\multicolumn{1}{r|}{MAE}&\multicolumn{1}{r}{MSE}&\multicolumn{1}{r|}{MAE}&\multicolumn{1}{r}{MSE}&\multicolumn{1}{r|}{MAE}&\multicolumn{1}{r}{MSE}&\multicolumn{1}{r|}{MAE}&\multicolumn{1}{r}{MSE}&\multicolumn{1}{r|}{MAE}&\multicolumn{1}{r}{MSE}&\multicolumn{1}{r|}{MAE}&\multicolumn{1}{r}{MSE}&\multicolumn{1}{r|}{MAE}&\multicolumn{1}{r}{MSE}&\multicolumn{1}{r|}{MAE}&\multicolumn{1}{r}{MSE}&\multicolumn{1}{r}{MAE}\\
    \midrule
    \multicolumn{1}{r|}{\multirow{5}{*}{\rotatebox{90}{ETTh1}}}&96&\multicolumn{1}{c}{\textbf{0.368}}&\textbf{0.391} &0.394 &0.409 &0.377 &\underline{0.396} &0.384 &0.402 & 0.386& 0.400&0.420 &0.439 &0.424 &0.432 &0.494 &0.479&\underline{0.376} &0.419\\
    \multicolumn{1}{r|}{}                                  &192&\multicolumn{1}{c}{\textbf{0.418}}&\textbf{0.422} &0.448 &0.440 & 0.425 &\underline{0.426} &0.436 &0.429 & 0.437&0.432 &0.541&0.520 &0.475 &0.462 &0.538 &0.504&\underline{0.420} &0.448\\
    \multicolumn{1}{r|}{}                                  &336&\multicolumn{1}{c}{\textbf{0.453}}&\textbf{0.440} &0.492 &0.465 & 0.461 &\underline{0.448} &0.491 &0.469  & 0.481&0.459 &0.722&0.648 &0.518 &0.488 &0.574 &0.521&\underline{0.459} &0.465\\
    \multicolumn{1}{r|}{}                                  &720&\multicolumn{1}{c}{\textbf{0.456}}&\textbf{0.458} &0.520 &0.503 & 0.529 &\underline{0.500} &0.521 &\underline{0.500} & 0.519&0.516 &0.811&0.691 &0.547 &0.533 &0.562 &0.535&\underline{0.506} &0.507\\
    \rowcolor{gray!40}\multicolumn{1}{r|}{}                                  &Avg&\multicolumn{1}{c}{\textbf{0.424}}&\textbf{0.428} &0.464 &0.454 &0.448 &\underline{0.443} &0.458 &0.450 & 0.456&0.452 &0.623 &0.574 &0.491 &0.479 &0.542 &0.510&\underline{0.440} &0.460\\
    \midrule
    \multicolumn{1}{r|}{\multirow{5}{*}{\rotatebox{90}{ETTh2}}}&96&\multicolumn{1}{c}{\textbf{0.283}}&\textbf{0.338} &\underline{0.297} &\underline{0.349} &0.310 &0.353 &0.340 &0.374 & 0.333&0.387 &0.745 &0.584 &0.397 &0.437 &0.340 &0.391&0.358 &0.397\\
    \multicolumn{1}{r|}{}                                  &192&\multicolumn{1}{c}{\textbf{0.365}}&\textbf{0.390} &\underline{0.380} &\underline{0.400} &0.390 &0.405 &0.402 &0.414 & 0.477&0.476 &0.877 &0.656 &0.520 &0.504 &0.430 &0.439&0.429 &0.439\\
    \multicolumn{1}{r|}{}                                  &336&\multicolumn{1}{c}{\textbf{0.399}}&\textbf{0.423} &\underline{0.428} &\underline{0.432} &0.430 &0.434 &0.452 &0.452 & 0.594&0.541 &1.043 &0.731 &0.626 &0.559 &0.485 &0.479&0.496 &0.487\\
    \multicolumn{1}{r|}{}                                  &720&\multicolumn{1}{c}{\textbf{0.420}}&\textbf{0.443} &\underline{0.427} &\underline{0.445} &0.438 &0.449 &0.462 &0.468 & 0.831&0.657 &1.104 &0.763 &0.863 &0.672 &0.500 &0.497 &0.463&0.474\\
    \rowcolor{gray!40}\multicolumn{1}{r|}{}                                  &Avg&\multicolumn{1}{c}{\textbf{0.367}}&\textbf{0.399} &\underline{0.383} &\underline{0.407} &0.392 &0.410 &0.414 &0.427 & 0.559&0.515 &0.942 &0.684 &0.602 &0.543 &0.439 &0.452&0.437 &0.449\\
    \midrule
    \multicolumn{1}{r|}{\multirow{5}{*}{\rotatebox{90}{ETTm1}}}&96&\multicolumn{1}{c}{\textbf{0.306}}&\textbf{0.358} &0.343 &0.378 &\underline{0.332} &\underline{0.369} &0.338 &0.375 &0.345 &0.372 &0.370 &0.404 &0.374 &0.400 &0.375 &0.398&0.379 &0.419\\
    \multicolumn{1}{r|}{}                                  &192&\multicolumn{1}{c}{\textbf{0.347}}&\textbf{0.380} &0.381 &0.395 &\underline{0.373} &0.389 &0.374 &\underline{0.387} & 0.380&0.389 &0.460 &0.488 &0.400 &0.407 &0.408 &0.410 &0.426&0.441\\
    \multicolumn{1}{r|}{}                                  &336&\multicolumn{1}{c}{\textbf{0.389}}&\textbf{0.401} &0.419 &0.418 &\underline{0.408} &\underline{0.411} &0.410 &\underline{0.411} & 0.413&0.413 &0.637 &0.607 &0.438 &0.438 &0.435 &0.428 &0.445&0.459\\
    \multicolumn{1}{r|}{}                                  &720&\multicolumn{1}{c}{\textbf{0.452}}&\textbf{0.434} &0.486 &0.456 &\underline{0.463} &\underline{0.446} &0.478 &0.450 & 0.474&0.453 &0.863 &0.720 &0.527 &0.502 &0.499 &0.462 &0.543&0.490\\
    \rowcolor{gray!40}\multicolumn{1}{r|}{}                                  &Avg&\multicolumn{1}{c}{\textbf{0.374}}&\textbf{0.393} &0.407 &0.411 &\underline{0.394} &\underline{0.404} &0.400 &0.406 & 0.403&0.407 &0.582 &0.555 &0.435 &0.437 &0.429 &0.425 &0.448&0.452\\
    \midrule
    \multicolumn{1}{r|}{\multirow{5}{*}{\rotatebox{90}{ETTm2}}}&96&\multicolumn{1}{c}{\textbf{0.170}}&\textbf{0.254} &0.183 &0.268 &\underline{0.177} &\underline{0.259} &0.187 &0.267 &0.193 &0.292 &0.270 &0.372 &0.209 &0.308 &0.189 &0.280&0.203 &0.287\\
    \multicolumn{1}{r|}{}                                  &192&\multicolumn{1}{c}{\textbf{0.231}}&\textbf{0.283} &0.252 &0.312 &0.242 &0.301 &0.249 &0.309 & 0.284&0.362 &0.242 &0.301 &0.311 &0.382 &0.253 &0.319 &0.269&0.328\\
    \multicolumn{1}{r|}{}                                  &336&\multicolumn{1}{c}{\textbf{0.296}}&\textbf{0.325} &0.313 &0.350 &0.303 &0.343 &0.321 &0.351 & 0.369&0.427 &0.303 &0.343 &0.442 &0.466 &0.314 &0.357 &0.325&0.366\\
    \multicolumn{1}{r|}{}                                  &720&\multicolumn{1}{c}{\textbf{0.405}}&\textbf{0.402} &0.411 &0.406 &\underline{0.410} &\underline{0.404} &0.408 &0.403 & 0.554&0.522 &0.410 &0.404 &0.675 &0.587 &0.414 &0.413 &0.421&0.415\\
    \rowcolor{gray!40}\multicolumn{1}{r|}{}                                  &Avg&\multicolumn{1}{c}{\textbf{0.276}}&\textbf{0.316} &0.290 &0.334 &0.283 &0.327 &0.291 &0.333 & 0.350&0.401 &0.306 &0.355 &0.409 &0.436 &0.293 &0.342 &0.305&0.349\\
    \midrule
    \multicolumn{1}{r|}{\multirow{5}{*}{\rotatebox{90}{Electricity}}}&96&\multicolumn{1}{c}{0.153}&0.259 &\textbf{0.148} &\textbf{0.240} &0.170 &0.260 &0.168 &0.272 &0.197 &0.282 &\underline{0.150} &\underline{0.253} &0.207 &0.307 &0.187 &0.304 &0.193&0.308\\
    \multicolumn{1}{r|}{}                                  &192&\multicolumn{1}{c}{\textbf{0.159}}&\textbf{0.245} &\underline{0.162} &\underline{0.253} &0.187 &0.276 &0.184 &0.289 & 0.196&0.285 &0.167 &0.267 &0.165 &0.264 &0.199 &0.315 &0.201&0.315\\
    \multicolumn{1}{r|}{}                                  &336&\multicolumn{1}{c}{\textbf{0.171}}&\underline{0.275} &\underline{0.178} &\textbf{0.269} &0.203 &0.291 &0.198 &0.300 & 0.209&0.301 &0.189 &0.287 &0.230 &0.333 &0.212 &0.329 &0.214&0.329\\
    \multicolumn{1}{r|}{}                                  &720&\multicolumn{1}{c}{\textbf{0.193}}&\textbf{0.287} &0.225 &0.317 &0.245 &0.325 &0.220 &0.320 & 0.245&0.333 &0.256 &0.337 &0.265 &0.360 &0.233 &0.345 &0.246&0.355\\
    \rowcolor{gray!40}\multicolumn{1}{r|}{}                                  &Avg&\multicolumn{1}{c}{\textbf{0.169}}&\textbf{0.267} &0.178 &0.270 &0.201 &0.288 &0.192 &0.295 & 0.212&0.300 &0.191 &0.286 &0.229 &0.329 &0.208 &0.323 &0.214&0.327\\
    \midrule
    \multicolumn{1}{r|}{\multirow{5}{*}{\rotatebox{90}{Weather}}}&96&\multicolumn{1}{c}{\textbf{0.168}}&\textbf{0.210} &0.174 &0.214 &0.179 &0.220 &\underline{0.172} &\underline{0.220} & 0.196&0.255 &0.174 &0.239 &0.182 &0.242 &0.197 &0.281 &0.217&0.296\\
    \multicolumn{1}{r|}{}                                  &192&\multicolumn{1}{c}{\textbf{0.196}}&\textbf{0.243} &0.221 &0.254 &0.222 &0.257 &\underline{0.219} &\underline{0.261} &0.237 &0.296 &0.221 &0.287 &0.227 &0.287 &0.237 &0.312 &0.276&0.336\\
    \multicolumn{1}{r|}{}                                  &336&\multicolumn{1}{c}{\textbf{0.261}}&\textbf{0.283} &0.278 &\underline{0.296} &0.279 &0.297 &0.280 &0.306 &0.283 &0.335 &\underline{0.277} &0.340 &0.282 &0.334 &0.298 &0.353 &0.339&0.380\\
    \multicolumn{1}{r|}{}                                  &720&\multicolumn{1}{c}{\underline{0.350}}&\textbf{0.345} &0.358 &\underline{0.349} &0.357 &0.351 &0.365 &0.359 &\textbf{0.345} &0.381 &0.371 &0.410 &0.352 &0.386 &0.352&0.288 &0.403&0.428\\
    \rowcolor{gray!40}\multicolumn{1}{r|}{}                                  &Avg&\multicolumn{1}{c}{\textbf{0.244}}&\textbf{0.270} &\underline{0.258} &\underline{0.279} &0.259 &0.281 &0.259 &0.287 & 0.265&0.317 &0.261 &0.319 &0.261 &0.312 &0.271 &0.334 &0.309&0.360\\
    \midrule
    \multicolumn{1}{r|}{\multirow{5}{*}{\rotatebox{90}{ILI}}}&24&\multicolumn{1}{c}{\textbf{2.003}}&\textbf{0.915} &3.014 &1.169 &2.281 &\underline{0.926} &2.317 &0.934 &\underline{2.215} &1.081 &3.478 &1.242 &8.313 &2.144 &2.527 &1.020 &3.228&1.260\\
    \multicolumn{1}{r|}{}                                  &36&\multicolumn{1}{c}{2.010}&\underline{0.923} &2.991 &1.172 &2.344 &0.938 &\underline{1.972} &\textbf{0.920} & \textbf{1.963}&0.963 &4.268 &1.399 &6.631 &1.902 &2.615 &1.007 &2.679&1.080\\
    \multicolumn{1}{r|}{}                                  &48&\multicolumn{1}{c}{\textbf{1.996}}&\textbf{0.910} &2.862 &1.146 &2.184 &\textbf{0.910} &2.238 &\underline{0.940} & \underline{2.130}&1.024 &3.774 &1.287 &7.299 &1.982 &2.359 &0.972 &2.622&1.078\\
    \multicolumn{1}{r|}{}                                  &60&\multicolumn{1}{c}{\textbf{1.978}}&\textbf{0.903} &2.955 &1.175 &2.092 &\underline{0.921} &\underline{2.027} &0.928 & 2.368&1.096 &4.009 &1.335 &7.283 &1.985 &2.487 &1.016 &2.857&1.157\\
    \rowcolor{gray!40}\multicolumn{1}{r|}{}                                  &Avg&\multicolumn{1}{c}{\textbf{1.997}}&\textbf{0.913} &2.956 &1.166 &2.225 &\underline{0.924} &\underline{2.139} &0.931 & 2.169&1.041 &3.882 &1.316 &7.382 &2.003 &2.497 &1.004 &2.847&1.144\\
    \midrule
    \multicolumn{1}{r|}{\multirow{5}{*}{\rotatebox{90}{Traffic}}}&96&\multicolumn{1}{c}{\underline{0.473}}&\underline{0.285} &\textbf{0.395} &\textbf{0.268} &0.544 &0.360 &0.593 &0.321 &0.650  &0.396 &0.522 &0.290 &0.615 &0.391 &0.607 &0.392&0.587 &0.366\\
    \multicolumn{1}{r|}{}                                  &192&\multicolumn{1}{c}{\underline{0.474}}&0.311 &\textbf{0.417} &\textbf{0.276} &0.540 &0.354 &0.617 &0.336 &0.598  &0.370 &0.530&\underline{0.293} &0.601 &0.382 &0.621 &0.399 &0.604&0.373\\
    \multicolumn{1}{r|}{}                                  &336&\multicolumn{1}{c}{\underline{0.495}}&\underline{0.300} &\textbf{0.433} &\textbf{0.283} &0.552 &0.360 &0.629 &0.336 &0.605 & 0.373&0.558 &0.305 &0.613 &0.386 &0.622 &0.396 &0.621&0.383\\
    \multicolumn{1}{r|}{}                                  &720&\multicolumn{1}{c}{\underline{0.528}}&\underline{0.313} &\textbf{0.467} &\textbf{0.302} &0.590 &0.378 &0.640 &0.350 &0.645 &0.394 &0.589 &0.328 &0.658 &0.407 &0.632 &0.396 &0.626&0.382\\
    \rowcolor{gray!40}\multicolumn{1}{r|}{}                                  &Avg&\multicolumn{1}{c}{\underline{0.493}}&\underline{0.302} &\textbf{0.428} &\textbf{0.282} &0.557 &0.363 &0.620 &0.336 & 0.625&0.383 &0.550 &0.304 &0.622 &0.392 &0.621 &0.396 &0.610&0.376\\
    \midrule
    \multicolumn{1}{r|}{\multirow{5}{*}{\rotatebox{90}{Exchange}}}&96&\multicolumn{1}{c}{\textbf{0.079}}&\textbf{0.197} &0.103 &0.231 &\underline{0.085} &\underline{0.202} &0.107 &0.234 &0.088  &0.218 &0.256 &0.367 &0.116 &0.262 &0.085 &0.204&0.148 &0.278\\
    \multicolumn{1}{r|}{}                                  &192&\multicolumn{1}{c}{\textbf{0.176}}&\textbf{0.294} &0.189 &0.313 &\underline{0.180} &\underline{0.301} &0.226 &0.334 &\textbf{0.176}  &0.315 &0.469&0.508 &0.182 &0.303 &0.271 &0.380 &0.604&0.373\\
    \multicolumn{1}{r|}{}                                  &336&\multicolumn{1}{c}{\textbf{0.297}}&\textbf{0.401} &0.377 &0.447 &0.336 &\underline{0.420} &0.367 &0.448 &\underline{0.313} & 0.427&0.975 &0.763 &0.377 &0.466 &0.348 &0.428 &0.460&0.500\\
    \multicolumn{1}{r|}{}                                  &720&\multicolumn{1}{c}{\textbf{0.823}}&\textbf{0.684} &0.870 &0.705 &0.881 &0.710 &0.964 &0.746 &0.839 &\underline{0.695} &1.618 &1.028 &\underline{0.831} &0.699 &1.025 &0.774 &1.195&0.841\\
    \rowcolor{gray!40}\multicolumn{1}{r|}{}                                  &Avg&\multicolumn{1}{c}{\textbf{0.343}}&\textbf{0.394} &0.384 &0.424 &0.371 &\underline{0.408} &0.416 &0.440 & \underline{0.354}&0.414 &0.830 &0.667 &0.377 &0.432 &0.432 &0.447 &0.602&0.498\\
    \bottomrule
    \end{tabular}
    \label{long}
\end{table*}
\subsection{Short-term forecasting}
\textbf{Setups} We also conduct short-term forecasting, which has a relatively short horizon, on M4 dataset \cite{M4}. The M4 dataset encompasses 100,000 distinct time series collected at various frequencies, categorized into 6 subsets: yearly, quarterly, monthly, weekly, daily, and hourly. The input lengths are set at twice the prediction lengths. Three metrics, namely SMAPE, MASE, and OWA, are employed for evaluation.\\
\indent \textbf{Results} Table \ref{short} demonstrates that our model is superior to other methods in most cases. Given that the time series in M4 dataset originate from diverse sources with varying properties, the remarkable performance of our Caformer underscores the robustness of the learned features.
\begin{table*}[htbp]
    % \fontsize{7.2}{8.8}\selectfont
    \scriptsize
    \renewcommand\arraystretch{1.15}
    \centering
    \caption{Short-term forecasting task. The prediction lengths are in $[6, 48]$. The results are averaged from several datasets under different sample intervals. Lower metrics indicate better performance. The best results are in \textbf{bold}  and the second best are \underline{underlined}.}
    \begin{tabular}{c|c|c|c|c|c|c|c|c|c|c}
    \toprule
    \multicolumn{2}{c|}{Models}&\multicolumn{1}{c|}{\cellcolor{gray!40}Ours}&
    \multicolumn{1}{c|}{\makecell[c]{iTransformer\\(2024)}}&\multicolumn{1}{c|}{\makecell[c]{PatchTST\\(2023)}}&\multicolumn{1}{c|}{\makecell[c]{TimesNet\\(2023)}}&\multicolumn{1}{c|}{\makecell[c]{DLinear\\(2023)}}&\multicolumn{1}{c|}{\makecell[c]{Crossformer\\(2023)}}&\multicolumn{1}{c|}{\makecell[c]{LightTS\\(2022)}}&\multicolumn{1}{c|}{\makecell[c]{ETSformer\\(2022)}}&\multicolumn{1}{c}{\makecell[c]{FEDformer\\(2022)}}\\
    \midrule
    \multicolumn{1}{r|}{\multirow{3}{*}{\rotatebox{90}{Yearly}}}&SMAPE&\multicolumn{1}{c|}{\cellcolor{gray!40}\textbf{13.365}}& 14.321&13.550 &\underline{13.387} &16.965 &13.392 &14.247 &18.009 &13.728 \\
    \multicolumn{1}{r|}{}                                  &MASE&\multicolumn{1}{c|}{\cellcolor{gray!40}\textbf{2.895}}&3.230 &3.028 & \underline{2.996}&4.283 &3.001 &3.109 &4.487 &3.048\\
    \multicolumn{1}{r|}{}                                  &OWA&\multicolumn{1}{c|}{\cellcolor{gray!40}\textbf{0.776}}&0.845 &0.796 &\underline{0.786} &0.781 &0.787 &0.827 &1.115 &0.803 \\
    \midrule
    \multicolumn{1}{r|}{\multirow{3}{*}{\rotatebox{90}{Quarterly}}}&SMAPE&\multicolumn{1}{c|}{\cellcolor{gray!40}\textbf{9.989}}&10.764 &10.195 &\underline{10.100} &12.145 &16.317 &11.364 & 13.376&10.792 \\
    \multicolumn{1}{r|}{}                                  &MASE&\multicolumn{1}{c|}{\cellcolor{gray!40}\textbf{1.171}}&1.284 &1.205 &\underline{1.182} &1.520 &2.197 &1.328 &1.906 &1.283\\
    \multicolumn{1}{r|}{}                                  &OWA&\multicolumn{1}{c|}{\cellcolor{gray!40}\textbf{0.882}}&0.956 &0.902 &\underline{0.890} &1.106 &1.542 &1.000 &1.302 &0.958 \\
    \midrule
    \multicolumn{1}{r|}{\multirow{3}{*}{\rotatebox{90}{Monthly}}}&SMAPE&\multicolumn{1}{c|}{\cellcolor{gray!40}\textbf{12.612}}&14.245 &12.96 &\underline{12.670} &13.514 &12.924 &14.014 &14.588 &14.260 \\
    \multicolumn{1}{r|}{}                                  &MASE&\multicolumn{1}{c|}{\cellcolor{gray!40}\textbf{0.914}}&1.110 &0.968 &\underline{0.933}&1.037 &0.966 &1.053 &1.368 &1.102\\
    \multicolumn{1}{r|}{}                                  &OWA&\multicolumn{1}{c|}{\cellcolor{gray!40}\textbf{0.861}}&1.016 &0.905 &\underline{0.878} &0.956 &0.902 &0.981&1.149 &1.012  \\
    \midrule
    \multicolumn{1}{r|}{\multirow{3}{*}{\rotatebox{90}{Others}}}&SMAPE&\multicolumn{1}{c|}{\cellcolor{gray!40}\textbf{4.723}}&5.780 &5.005 &\underline{4.891} &6.709 &5.493&15.880&7.267 &4.954  \\
    \multicolumn{1}{r|}{}                                  &MASE&\multicolumn{1}{c|}{\cellcolor{gray!40}\textbf{3.105}}&4.153 &3.394 &\underline{3.302} &4.953 &3.690 &3.690&5.240 &3.264  \\
    \multicolumn{1}{r|}{}                                  &OWA&\multicolumn{1}{c|}{\cellcolor{gray!40}\textbf{0.991}}&1.263 &1.062 &\underline{1.035} &1.487 &1.160 &3.474&1.591 &1.036 \\
    \midrule
    \multicolumn{1}{r|}{\multirow{3}{*}{\rotatebox{90}{W.Avg.}}}&SMAPE&\multicolumn{1}{c|}{\cellcolor{gray!40}\textbf{11.734}}&12.999 &12.034 &11.829 &13.639 &13.474 &13.525&14.718&12.840  \\
    \multicolumn{1}{r|}{}                                  &MASE&\multicolumn{1}{c|}{\cellcolor{gray!40}\textbf{1.571}}&1.792 &1.620 &\underline{1.585} &2.095 &1.866 &2.111 &2.408 &1.701\\
    \multicolumn{1}{r|}{}                                  &OWA&\multicolumn{1}{c|}{\cellcolor{gray!40}\textbf{0.842}}&0.948 &0.867 &\underline{0.851} &1.051 &0.985 &1.051 &1.172 &0.918 \\
    \bottomrule
    \end{tabular}
    \label{short}
\end{table*}
\subsection{Imputation}
\textbf{Setups} The absence of data significantly hampers the performance of downstream analysis tasks, and time series imputation is a common practice employed to address missing data resulting from malfunctions. We select ETT \cite{informer}, Electricity\footref{elefoot}, and Weather\footref{weafoot} as our benchmarks. The time points are randomly masked in the ratio of \{12.5\%, 25\%, 37.5\%, 50\%\} to imitate different proportions of missing data. Both the MSE and MAE are used as metrics.\\
\indent \textbf{Results} Table \ref{imputation} shows the great performance of our model in imputation tasks. We argue that the insufficiency of data has a certain impact on the learning of cross-dimension and cross-time dependencies, thereby introducing undesirable biases. Thus, some outcomes in Table \ref{imputation} are not particularly favorable. Addressing these challenges constitutes a focal point for our future research direction.
\subsection{Classification}
\begin{figure}[ht]
    \centering
    % \vspace{-0.3cm}
    \includegraphics[height=5.5cm]{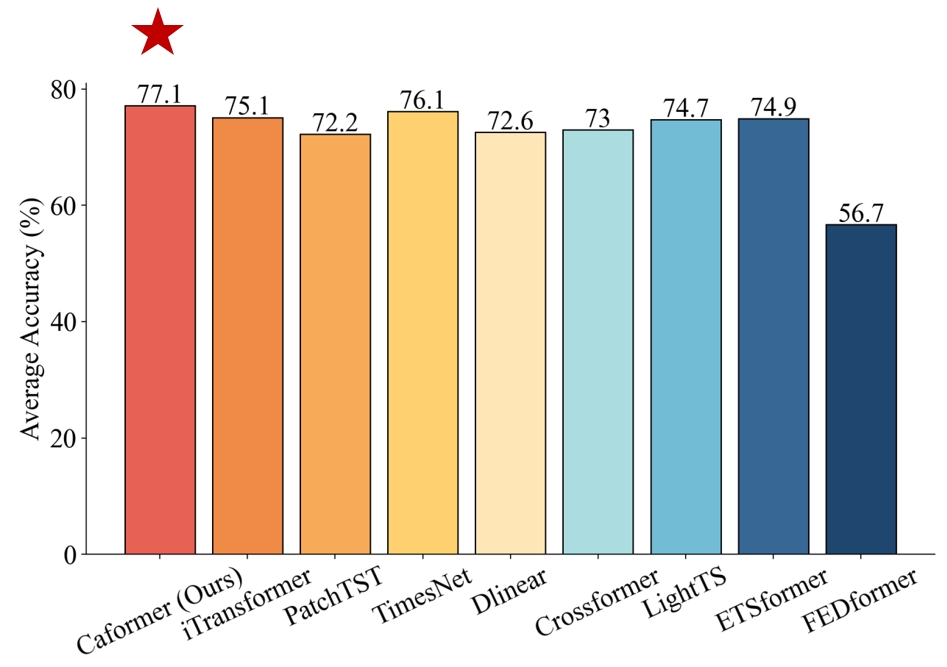}
    \caption{Model comparison in classification. The results are averaged from 10 subsets of UEA.}
    \label{Classification fig}
\end{figure}

\textbf{Setups} Time series classification aims to distinguish meaningful data patterns and effectively classify each time series. We conduct the experiments with 10 datasets from UEA Time Series Classification Archive \cite{UEA}. The classification accuracy is used as the metric. \\
\indent \textbf{Results} As shown in Fig. \ref{Classification fig}, Caformer achieves the best performance with an average accuracy of 77.1\%. By emphasizing dynamic interactions within time series, our model excels in its ability to acquire high-level representations.
\subsection{Anomaly detection}
\textbf{Setups} Anomaly detection aims to uncover anomalies in time series. We select five datasets, including SMD \cite{SMD}, MSL \cite{MSL}, SMAP \cite{MSL}, SWaT \cite{SWaT}, and PSM \cite{PSM}, as benchmarks.\\
\indent \textbf{Results} Table \ref{anomaly} demonstrates the excellent performance of Caformer in anomaly detection. The existence of anomalies introduces biases during inference, just as the problem in imputation tasks. Subsequent research will be undertaken to address this issue.
\begin{table*}[htbp]
    % \fontsize{7.2}{8.8}\selectfont
    \scriptsize
    \renewcommand\arraystretch{1.15}
    \setlength{\tabcolsep}{5pt}
    \centering
    \caption{Imputation task. We randomly mask \{12.5\%, 25\%, 37.5\%, 50\%\} of time points in the length-96 time series. The results are averaged from 4 different mask ratios. A lower MSE or MAE indicates a better performance. The best results are in \textbf{bold} and the second best are \underline{underlined}.}
    \begin{tabular}{c|c|cc|cc|cc|cc|cc|cc|cc|cc|cc}
    \toprule
    \multicolumn{2}{c|}{Models}&\multicolumn{2}{c|}{Ours}&\multicolumn{2}{c|}{\makecell[c]{iTransformer\\(2024)}}&\multicolumn{2}{c|}{\makecell[c]{PatchTST\\(2023)}}&\multicolumn{2}{c|}{\makecell[c]{TimesNet\\(2023)}}&\multicolumn{2}{c|}{\makecell[c]{DLinear\\(2023)}}&\multicolumn{2}{c|}{\makecell[c]{Crossformer\\(2023)}}&\multicolumn{2}{c|}{\makecell[c]{LightTS\\(2022)}}&\multicolumn{2}{c|}{\makecell[c]{ETSformer\\(2022)}}&\multicolumn{2}{c}{\makecell[c]{FEDformer\\(2022)}}\\
    \midrule
    \multicolumn{2}{c|}{Metric}&\multicolumn{1}{c}{MSE}&\multicolumn{1}{c|}{MAE}&\multicolumn{1}{c}{MSE}&\multicolumn{1}{c|}{MAE}&\multicolumn{1}{c}{MSE}&\multicolumn{1}{c|}{MAE}&\multicolumn{1}{c}{MSE}&\multicolumn{1}{c|}{MAE}&\multicolumn{1}{c}{MSE}&\multicolumn{1}{c|}{MAE}&\multicolumn{1}{c}{MSE}&\multicolumn{1}{c|}{MAE}&\multicolumn{1}{c}{MSE}&\multicolumn{1}{c|}{MAE}&\multicolumn{1}{c}{MSE}&\multicolumn{1}{c|}{MAE}&\multicolumn{1}{c}{MSE}&\multicolumn{1}{c}{MAE}\\
    \midrule
    \multicolumn{1}{r|}{\multirow{5}{*}{\rotatebox{90}{ETTh1}}}&12.5\%&\multicolumn{1}{c}{0.073}&\underline{0.168}&0.097&0.219&0.091&0.199&\textbf{0.057}&\textbf{0.159}&0.151&0.267&0.170&0.284&0.240&0.345&0.126&0.263&\underline{0.070}&0.190\\
    \multicolumn{1}{r|}{}                                  &25\%&\multicolumn{1}{c}{\underline{0.083}}&\underline{0.192}&0.125&0.249 &0.105&0.215&\textbf{0.069}&\textbf{0.178}&0.180&0.292&0.188&0.300&0.265&0.364&0.169&0.304&0.106&0.236\\
    \multicolumn{1}{r|}{}                                  &37.5\%&\multicolumn{1}{c}{\underline{0.104}}&\underline{0.211}&0.158&0.280 &0.122&0.232&\textbf{0.084}&\textbf{0.196}&0.215&0.318&0.210&0.318&0.296&0.382&0.220&0.347&0.124&0.258\\
    \multicolumn{1}{r|}{}                                  &50\%&\multicolumn{1}{c}{\underline{0.117}}&\underline{0.223}&0.225&0.338 &0.143&0.248&\textbf{0.102}&\textbf{0.215}&0.257&0.347&0.243&0.342&0.334&0.404&0.293&0.402&0.165&0.299\\
    \rowcolor{gray!40}\multicolumn{1}{r|}{}                                  &Avg&\multicolumn{1}{c}{\underline{0.094}}&\underline{0.198}&0.151&0.272 &0.115&0.224&\textbf{0.078}&\textbf{0.187}&0.201&0.306&0.203&0.311&0.284&0.373&0.202&0.329&0.117&0.246\\
    \midrule
    \multicolumn{1}{r|}{\multirow{5}{*}{\rotatebox{90}{ETTh2}}}&12.5\%&\multicolumn{1}{c}{\underline{0.050}}&\underline{0.156}&0.094&0.209 &0.055&0.149&\textbf{0.040}&\textbf{0.130}&0.100&0.216&0.135&0.242&0.101&0.231&0.187&0.319&0.095&0.212\\
    \multicolumn{1}{r|}{}                                  &25\%&\multicolumn{1}{c}{\underline{0.056}}&\underline{0.147}&0.119&0.237 &0.061&0.158&\textbf{0.046}&\textbf{0.141}&0.127&0.247&0.158&0.262&0.115&0.246&0.279&0.390&0.137&0.258\\
    \multicolumn{1}{r|}{}                                  &37.5\%&\multicolumn{1}{c}{\underline{0.061}}&\underline{0.154}&0.148&0.265 &0.066&0.166&\textbf{0.052}&\textbf{0.151}&0.158&0.276&0.176&0.279&0.126&0.257&0.400&0.465&0.187&0.304\\
    \multicolumn{1}{r|}{}                                  &50\%&\multicolumn{1}{c}{\underline{0.065}}&\underline{0.161}&0.194&0.303 &0.073&0.174&\textbf{0.060}&\textbf{0.162}&0.183&0.299&0.208&0.306&0.136&0.268&0.602&0.572&0.232&0.341\\
    \rowcolor{gray!40}\multicolumn{1}{r|}{}                                  &Avg&\multicolumn{1}{c}{\underline{0.058}}&\underline{0.155}&0.139&0.254 &0.064&0.162&\textbf{0.049}&\textbf{0.146}&0.142&0.259&0.169&0.272&0.119&0.250&0.367&0.436&0.163&0.279\\
    \midrule
    \multicolumn{1}{r|}{\multirow{5}{*}{\rotatebox{90}{ETTm1}}}&12.5\%&\multicolumn{1}{c}{\textbf{0.013}}&\textbf{0.081}&0.045&0.147 &\underline{0.019}&0.147&\underline{0.019}&\underline{0.092}&0.058&0.162&0.051&0.158&0.075&0.180&0.067&0.188&0.035&0.135\\
    \multicolumn{1}{r|}{}                                  &25\%&\multicolumn{1}{c}{\textbf{0.018}}&\textbf{0.091}&0.060&0.171 &0.060&0.171&\underline{0.023}&\underline{0.101}&0.080&0.193&0.048&0.141&0.093&0.206&0.096&0.229&0.052&0.166\\
    \multicolumn{1}{r|}{}                                  &37.5\%&\multicolumn{1}{c}{\textbf{0.022}}&\textbf{0.099}&0.077&0.195 &0.076&0.194&\underline{0.029}&\underline{0.111}&0.103&0.219&0.059&0.170&0.113&0.231&0.133&0.271&0.069&0.191\\
    \multicolumn{1}{r|}{}                                  &50\%&\multicolumn{1}{c}{\textbf{0.028}}&\textbf{0.114}&0.104&0.229 &0.102&0.226&\underline{0.036}&\underline{0.124}&0.132&0.248&0.067&0.181&0.134&0.255&0.186&0.323&0.089&0.218\\
    \rowcolor{gray!40}\multicolumn{1}{r|}{}                                  &Avg&\multicolumn{1}{c}{\textbf{0.020}}&\textbf{0.096}&0.072&0.186 &0.064&0.185&\underline{0.027}&\underline{0.107}&0.093&0.206&0.056&0.163&0.104&0.218&0.120&0.253&0.062&0.177\\
    \midrule
    \multicolumn{1}{r|}{\multirow{5}{*}{\rotatebox{90}{ETTm2}}}&12.5\%&\multicolumn{1}{c}{\underline{0.025}}&\underline{0.093}&0.052&0.152 &0.051&0.151&\textbf{0.018}&\textbf{0.080}&0.062&0.166&0.025&0.092&0.034&0.127&0.108&0.239&0.056&0.159\\
    \multicolumn{1}{r|}{}                                  &25\%&\multicolumn{1}{c}{\underline{0.027}}&\underline{0.098}&0.070&0.179 &0.124&0.248&\textbf{0.020}&\textbf{0.085}&0.085&0.196&0.086&0.193&0.042&0.143&0.164&0.294&0.080&0.195\\
    \multicolumn{1}{r|}{}                                  &37.5\%&\multicolumn{1}{c}{\underline{0.030}}&\underline{0.104}&0.091&0.203 &0.157&0.280&\textbf{0.023}&\textbf{0.091}&0.106&0.222&0.091&0.198&0.051&0.159&0.237&0.356&0.110&0.231\\
    \multicolumn{1}{r|}{}                                  &50\%&\multicolumn{1}{c}{\underline{0.033}}&\underline{0.110}&0.116&0.231 &0.214&0.329&\textbf{0.026}&\textbf{0.098}&0.131&0.247&0.097&0.204&0.059&0.174&0.323&0.421&0.156&0.276\\
    \rowcolor{gray!40}\multicolumn{1}{r|}{}                                  &Avg&\multicolumn{1}{c}{\underline{0.029}}&\underline{0.101}&0.082&0.191 &0.137&0.252&\textbf{0.022}&\textbf{0.088}&0.096&0.208&0.075&0.172&0.046&0.151&0.208&0.327&0.101&0.215\\
    \midrule
    \multicolumn{1}{r|}{\multirow{5}{*}{\rotatebox{90}{Electricity}}}&12.5\%&\multicolumn{1}{c}{\textbf{0.058}}&\textbf{0.167}&0.073&0.190 &\underline{0.072}&\underline{0.189}&0.085&0.202&0.092&0.214&0.079&0.199&0.102&0.229&0.196&0.321&0.107&0.237\\
    \multicolumn{1}{r|}{}                                  &25\%&\multicolumn{1}{c}{\textbf{0.068}}&\textbf{0.182}&0.090&0.214 &0.090&\underline{0.203}&\underline{0.089}&0.206&0.118&0.247&0.092&0.217&0.121&0.252&0.207&0.332&0.120&0.251\\
    \multicolumn{1}{r|}{}                                  &37.5\%&\multicolumn{1}{c}{\textbf{0.080}}&\textbf{0.197}&0.106&0.234 &0.106&0.234&\underline{0.094}&\underline{0.213}&0.144&0.276&0.103&0.230&0.141&0.273&0.219&0.344&0.136&0.266\\
    \multicolumn{1}{r|}{}                                  &50\%&\multicolumn{1}{c}{\textbf{0.097}}&\textbf{0.210}&0.126&0.256 &0.127&0.257&\underline{0.100}&\underline{0.221}&0.175&0.305&0.115&0.244&0.160&0.293&0.235&0.357&0.158&0.284\\
    \rowcolor{gray!40}\multicolumn{1}{r|}{}                                  &Avg&\multicolumn{1}{c}{\textbf{0.076}}&\textbf{0.189}&0.988&0.224 &0.099&0.221&\underline{0.092}&\underline{0.210}&0.132&0.260&0.097&0.223&0.131&0.262&0.214&0.339&0.130&0.259\\
    \midrule
    \multicolumn{1}{r|}{\multirow{5}{*}{\rotatebox{90}{Weather}}}&12.5\%&\multicolumn{1}{c}{\textbf{0.020}}&\textbf{0.033}&0.038&0.086 &0.029&0.049&\underline{0.025}&\underline{0.045}&0.039&0.084&0.039&0.095&0.047&0.101&0.057&0.141&0.041&0.107\\
    \multicolumn{1}{r|}{}                                  &25\%&\multicolumn{1}{c}{\textbf{0.025}}&\textbf{0.041}&0.046&0.105 &0.031&0.053&\underline{0.029}&\underline{0.052}&0.048&0.103&0.042&0.102&0.052&0.111&0.065&0.155&0.064&0.163\\
    \multicolumn{1}{r|}{}                                  &37.5\%&\multicolumn{1}{c}{\textbf{0.028}}&\textbf{0.047}&0.056&0.123 &0.034&0.058&\underline{0.031}&\underline{0.057}&0.057&0.121&0.045&0.106&0.058&0.121&0.081&0.180&0.107&0.229\\
    \multicolumn{1}{r|}{}                                  &50\%&\multicolumn{1}{c}{\textbf{0.031}}&\textbf{0.053}&0.069&0.143 &0.039&0.066&\underline{0.034}&\underline{0.062}&0.066&0.134&0.047&0.112&0.065&0.133&0.102&0.207&0.183&0.312\\
    \rowcolor{gray!40}\multicolumn{1}{r|}{}                                  &Avg&\multicolumn{1}{c}{\textbf{0.026}}&\textbf{0.043}&0.052&0.114 &0.033&0.057&\underline{0.030}&\underline{0.054}&0.052&0.110&0.043&0.104&0.055&0.117&0.076&0.171&0.099&0.203\\
    \bottomrule
    \end{tabular}
    \label{imputation}
\end{table*}
\begin{table*}[htbp]
    % \fontsize{7.2}{8.8}\selectfont
    \scriptsize
    \renewcommand\arraystretch{1.15}
    \centering
    \caption{Anomaly detection task. A higher value of Precision, Recall, and F1-score indicates a better performance. The best results are in \textbf{bold} and the second best are \underline{underlined}.}
    \begin{tabular}{c|c|c|c|c|c|c|c|c|c|c}
    \toprule
    \multicolumn{2}{c|}{Models}&\multicolumn{1}{c|}{Ours}&\multicolumn{1}{c|}{\makecell[c]{iTransformer\\(2024)}}&\multicolumn{1}{c|}{\makecell[c]{PatchTST\\(2023)}}&\multicolumn{1}{c|}{\makecell[c]{TimesNet\\(2023)}}&\multicolumn{1}{c|}{\makecell[c]{DLinear\\(2023)}}&\multicolumn{1}{c|}{\makecell[c]{Crossformer\\(2023)}}&\multicolumn{1}{c|}{\makecell[c]{LightTS\\(2022)}}&\multicolumn{1}{c|}{\makecell[c]{ETSformer\\(2022)}}&\multicolumn{1}{c}{\makecell[c]{FEDformer\\(2022)}}\\
    \midrule
    \multicolumn{1}{r|}{\multirow{3}{*}{SMD}}&Precision&\multicolumn{1}{c|}{90.21}&76.28 &87.5 &88.66 &83.62 & 87.44&87.10 &87.44 &87.95 \\
    \multicolumn{1}{r|}{}                                  &Recall&\multicolumn{1}{c|}{83.18}&85.57 &82.2 &83.14 &71.52 &59.10 &78.42 &79.23 &82.39\\
    \multicolumn{1}{r|}{}                                  &F1-score&\multicolumn{1}{c|}{\textbf{86.55}}&80.66 &84.7 &\underline{85.81} &77.10 &70.53 &82.53 &83.13 &85.08 \\
    \midrule
    \multicolumn{1}{r|}{\multirow{3}{*}{MSL}}&Precision&\multicolumn{1}{c|}{89.58}&86.15 &88.34 &83.92 &84.34 &90.33 &82.40 &85.13 &77.14 \\
    \multicolumn{1}{r|}{}                                  &Recall&\multicolumn{1}{c|}{78.56}&62.63 &70.92 &86.42 &85.42 &72.78 &75.78 &84.93 &80.07\\
    \multicolumn{1}{r|}{}                                  &F1-score&\multicolumn{1}{c|}{83.71}&72.53 &78.68 &\textbf{85.15} &\underline{84.88} &80.69 &78.95 &85.03 &78.57 \\
    \midrule
    \multicolumn{1}{r|}{\multirow{3}{*}{SMAP}}&Precision&\multicolumn{1}{c|}{92.54}&90.68 &90.41 &92.52 &92.32 & 89.91&92.58 &92.25 &90.47 \\
    \multicolumn{1}{r|}{}                                  &Recall&\multicolumn{1}{c|}{62.45}&52.77 &55.61 &58.29 &55.41 &55.42 &55.27 &55.75 &58.10\\
    \multicolumn{1}{r|}{}                                  &F1-score&\multicolumn{1}{c|}{\textbf{74.57}}&66.71 &68.86 &\underline{71.52} &69.26 & 68.57&69.21 &69.50 &70.76 \\
    \midrule
    \multicolumn{1}{r|}{\multirow{3}{*}{SWAT}}&Precision&\multicolumn{1}{c|}{94.65}&92.21 &90.95 &86.76 &80.91 &97.87 &91.98 &90.02 &90.17 \\
    \multicolumn{1}{r|}{}                                  &Recall&\multicolumn{1}{c|}{89.36}&93.08 &79.74 &97.32 &95.30 &84.44&94.72 &80.36 &96.42\\
    \multicolumn{1}{r|}{}                                  &F1-score&\multicolumn{1}{c|}{91.93}&92.64 &84.98 &91.74 &87.52& 90.66&\textbf{93.33} &84.91 &\underline{93.19} \\
    \midrule
    \multicolumn{1}{r|}{\multirow{3}{*}{PSM}}&Precision&\multicolumn{1}{c|}{98.82}&97.96 &98.51 &98.19 &98.28 & 98.64 &98.37 &99.31 &97.31 \\
    \multicolumn{1}{r|}{}                                  &Recall&\multicolumn{1}{c|}{95.41}&92.10 &93.44 &96.76 &89.26 &94.66 &95.97 &85.28 &97.16\\
    \multicolumn{1}{r|}{}                                  &F1-score&\multicolumn{1}{c|}{\underline{97.09}}&94.94 &96.11 &97.47 &93.55 &96.61 &97.15 &91.76 &\textbf{97.23} \\
    \midrule
    \rowcolor{gray!40}\multicolumn{2}{c|}{Average F1-score}&
    \textbf{86.77}&81.49 &82.66 &\underline{86.34} &82.46 &81.41 &84.23 &82.87 &84.97 \\
    \bottomrule
    \end{tabular}
    \label{anomaly}
\end{table*}
\begin{table*}[!ht]
    \scriptsize
    \centering
    \renewcommand\arraystretch{1.15}
    \setlength{\tabcolsep}{5pt}
    \caption{Ablation of different components in Caformer in long-term forecasting task. The results are averaged from different input lengths.}
    \begin{tabular}{c|cc|cc|cc|cc|cc|cc|cc|cc|cc}
        \toprule
         \multicolumn{1}{c|}{Dataset}&\multicolumn{2}{c|}{ETTh1}&\multicolumn{2}{c|}{ETTh2}&\multicolumn{2}{c|}{ETTm1}&\multicolumn{2}{c|}{ETTm2}&\multicolumn{2}{c|}{Electricity}&\multicolumn{2}{c|}{Weather}&\multicolumn{2}{c|}{ILI}&\multicolumn{2}{c|}{Traffic}&\multicolumn{2}{c}{Exchange}  \\
         \midrule
         \multicolumn{1}{c|}{Metric}&\multicolumn{1}{c}{MSE}&\multicolumn{1}{r|}{MAE}&\multicolumn{1}{r}
         {MSE}&\multicolumn{1}{r|}{MAE}&\multicolumn{1}{r}{MSE}&\multicolumn{1}{r|}{MAE}&\multicolumn{1}{r}{MSE}&\multicolumn{1}{r|}{MAE}&\multicolumn{1}{r}{MSE}&\multicolumn{1}{r|}{MAE}&\multicolumn{1}{r}{MSE}&\multicolumn{1}{r|}{MAE}&\multicolumn{1}{r}{MSE}&\multicolumn{1}{r|}{MAE}&\multicolumn{1}{r}{MSE}&\multicolumn{1}{r|}{MAE}&\multicolumn{1}{r}{MSE}&\multicolumn{1}{r}{MAE}\\
         \midrule
         w/o Dep&0.443&0.438&0.383&0.406&0.386&0.401&0.281&0.324&0.199&0.283&0.251&0.278&2.210&0.916&0.536&0.356&0.367&0.400\\
         w/o Dyn&0.431&0.437&0.375&0.409&0.383&0.409&0.291&0.324&0.175&0.312&0.265&0.283&2.145&0.920&0.532&0.348&0.365&0.401\\
         w/o Env&0.436&0.438&0.378&0.405&0.385&0.396&0.281&0.324&0.198&0.282&0.251&0.278&2.116&0.917&0.551&0.358&0.367&0.401\\
         
         \rowcolor{gray!40} Caformer&\textbf{0.424}&\textbf{0.428}&\textbf{0.367}&\textbf{0.399}&\textbf{0.374}&\textbf{0.393}&\textbf{0.276}&\textbf{0.316}&\textbf{0.169}&\textbf{0.267}&\textbf{0.244}&\textbf{0.270}&\textbf{1.997}&\textbf{0.913}&\textbf{0.493}&\textbf{0.302}&\textbf{0.343}&\textbf{0.394}\\
         \bottomrule
    \end{tabular}
    \label{ablation}
\end{table*} 
\subsection{Ablation study and Hyper-parameter sensitivity}
\subsubsection{Ablation study}
To examine the effectiveness of each component in Caformer, we perform ablation experiments with the following variants: a) \textbf{w/o Dep}, which omits Dependency Learner for interactions among dependencies. b) \textbf{w/o Env}, which excludes Dynamic Learner for environment information. c) \textbf{w/o Dyn}, which does not utilize Dynamic Learner for dynamic interactions related to cross-dimension dependency. The results are shown in Table \ref{ablation}.
\subsubsection{Hyper-parameter sensitivity}
\begin{figure}
    \centering
    \includegraphics[height=4.5cm]{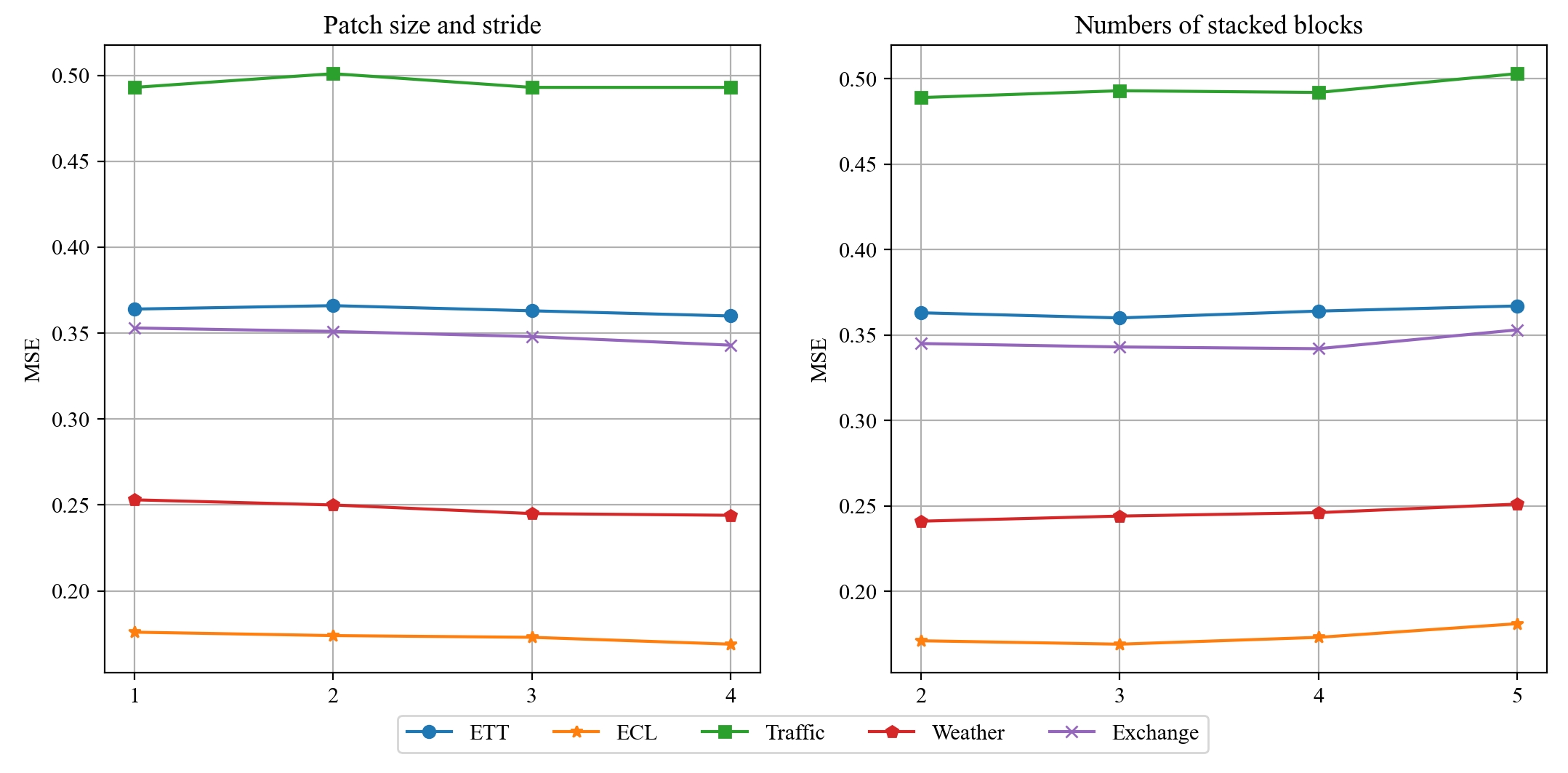}
    \caption{Hyperparameter sensitivity with respect to the patch size and stride and the numbers of stacked block.}
    \label{Appendix-hyper}
\end{figure}

\begin{figure}
    \centering
    \includegraphics[height=4.5cm]{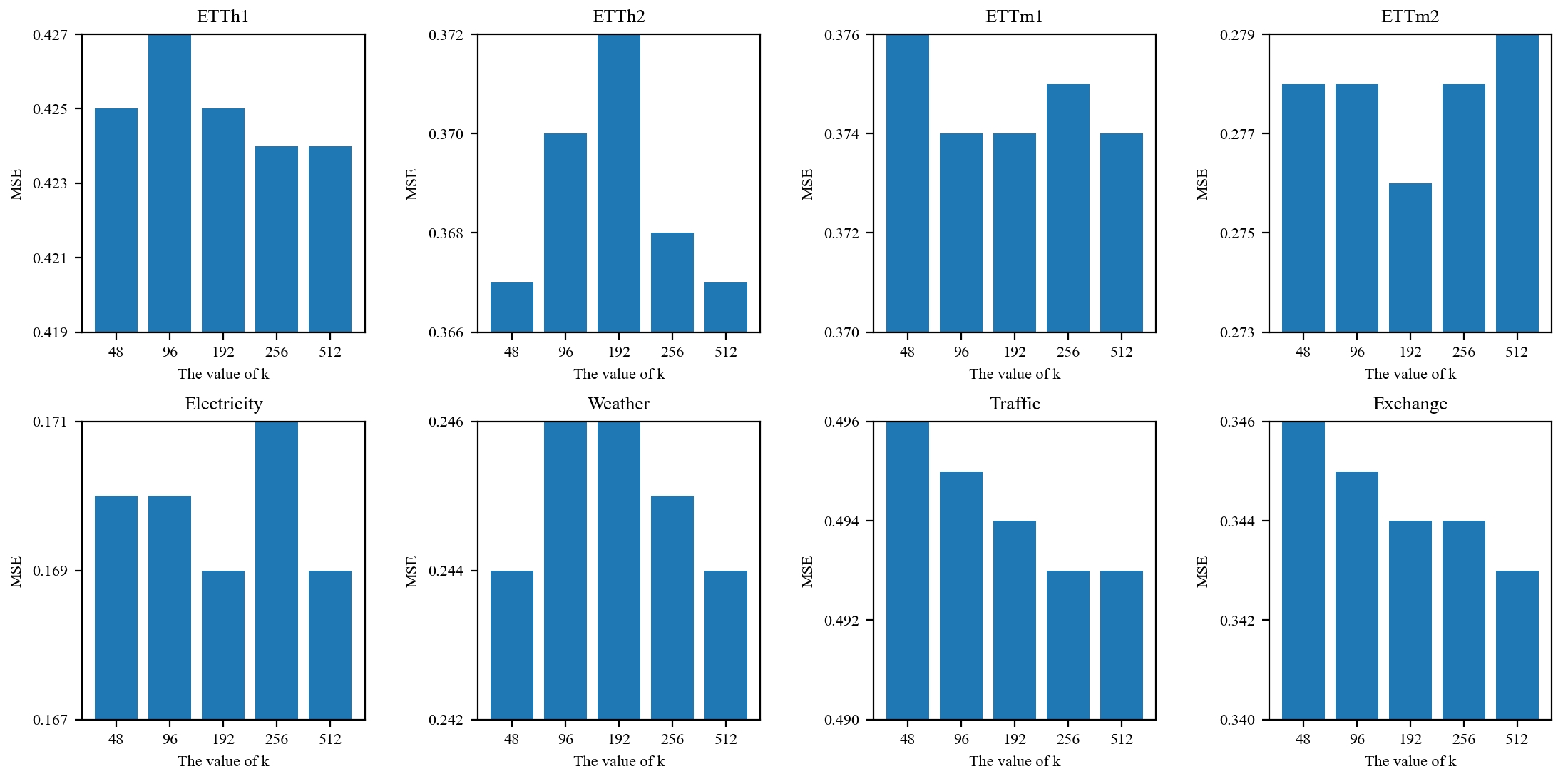}
    \caption{Hyper-parameter sensitivity with respect to the shape of environment aligning matrix.}
    \label{attention-block}
\end{figure}

To assess the hyperparameter sensitivity of Caformer, we perform experiments with varying model parameters, including the patch size and stride, the number of stacked blocks, and the shape of the aligning matrix. The detailed results are provided below. \\
\textbf{Results of different patch sizes and strides.} To verify the impact of the patch size and the stride, we perform experiments with four settings: (P = 4, S = 4), (P = 8, S = 4), (P = 8, S = 8), and (P = 16, S = 8). Results are shown in Fig. \ref{Appendix-hyper}. It can be inferred that the performance is stable for different settings of the patch size and the stride, indicating the robustness of our model to these two parameters. Given that the optimal patch size and stride may vary across datasets, we conduct the setting as (P = 16, S = 8) in our model.\\
\textbf{Results of different numbers of stacked blocks.} To verify the model's sensitivity to different numbers of blocks, the numbers of stacked blocks are selected as 2, 3, 4, and 5 to make the comparison. Results can be found in Fig. \ref{Appendix-hyper}. It can be inferred that the number of stacked blocks has an effect on performance, and it is not favored to be large concerning efficiency and performance. As a result, the number of stacked blocks is selected as 3 in our model.\\
\textbf{Results of different shapes of aligning matrices.} As stated above, $\textbf{H}_{e}\in \mathbb{R}^{k\times k}$ is used as the environment aligning matrix to fuse the environmental factors, and its shape $k$ is the predefined hyper-parameter determining the stratifying condition. To verify the impact of the shape $k$, we conduct experiments with $k \in \{48,96,192,256,512\}$. Results are shown in Fig. \ref{attention-block}. It can be inferred that the performance doesn't vary significantly with different shapes of aligning matrix. This finding indicates the robustness of our model to this parameter. Moreover, it's worth noting that the shapes of aligning matrices may be influenced by the characteristics of the corresponding dataset. As a general recommendation, a shape of 256 for aligning matrices is considered a good choice for most datasets.
\begin{figure}
    \centering
    \includegraphics[height=5cm]{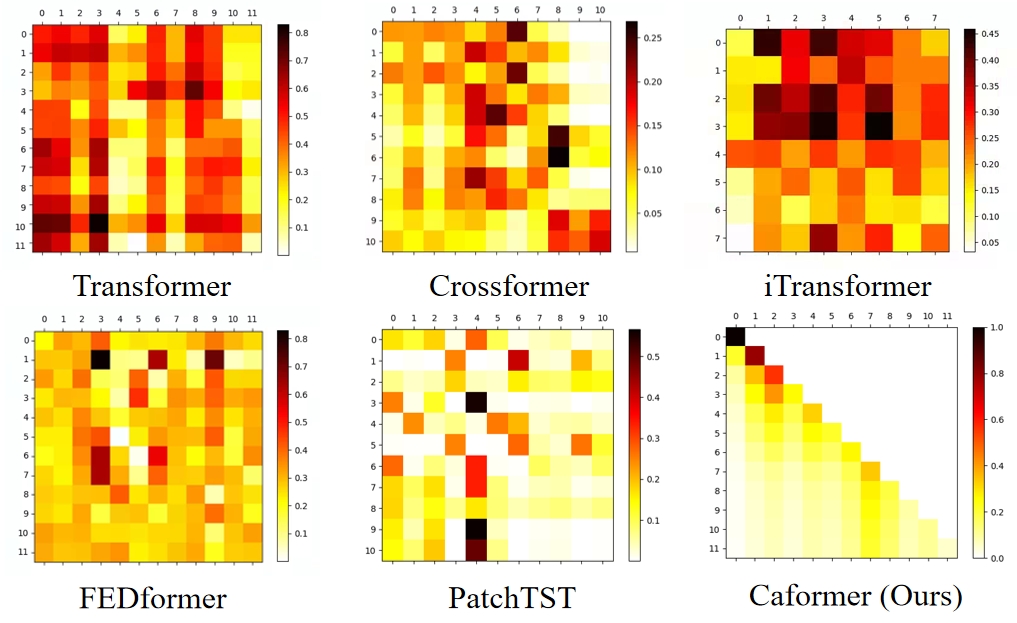}
    \caption{Comparison of attention maps or alignment matrix obtained by different models. These visualizations of correlations within time series indicate models' capacity in capturing interactions.}
    % \vspace{-0.5cm}
    \label{attention}
\end{figure}
\subsection{Interpretability analysis}
Transformer-based methods leverage the self-attention mechanism, facilitating the learning of long-range dependencies via query-key-value dot product attention. These dot product operations ascertain the significance of a particular token in relation to others, fostering information interactions and alignment between tokens. However, as illustrated in Fig. \ref{attention}, attention maps inferred by prior methods often lack interpretability. This arises from the inherent nature of these methods, which learn alignment rules based on correlation, disregarding the fundamental causality. Indeed, self-attention mimics a content-based retrieval process, utilizing pairwise interactions grounded in statistical correlation. To enhance the comparability of individual methods in correlation learning, we standardize the alignment matrix to a value of 12 using the ETTh1 dataset. This is because Caformer computes the relationship between 12 time periods that can be considered to have the same underlying causal structure. For matrices exceeding dimensions of $12\times12$, we patch the original matrix and compute the mean value for each patch, representing the relationship between the region and its neighbors. As depicted in Fig. \ref{attention}, Caformer exhibits the strongest correlation along the matrix diagonal, and this correlation is not uniformly 1. This consistency with causality elucidates the impact of external effects. Also, the time restriction map effectively guides learned alignments to adhere to Granger causality, ensuring causes precede their effects, resulting in a lower-triangular matrix.
\begin{figure*}[htbp]
    \centering
    \includegraphics[height=5cm]{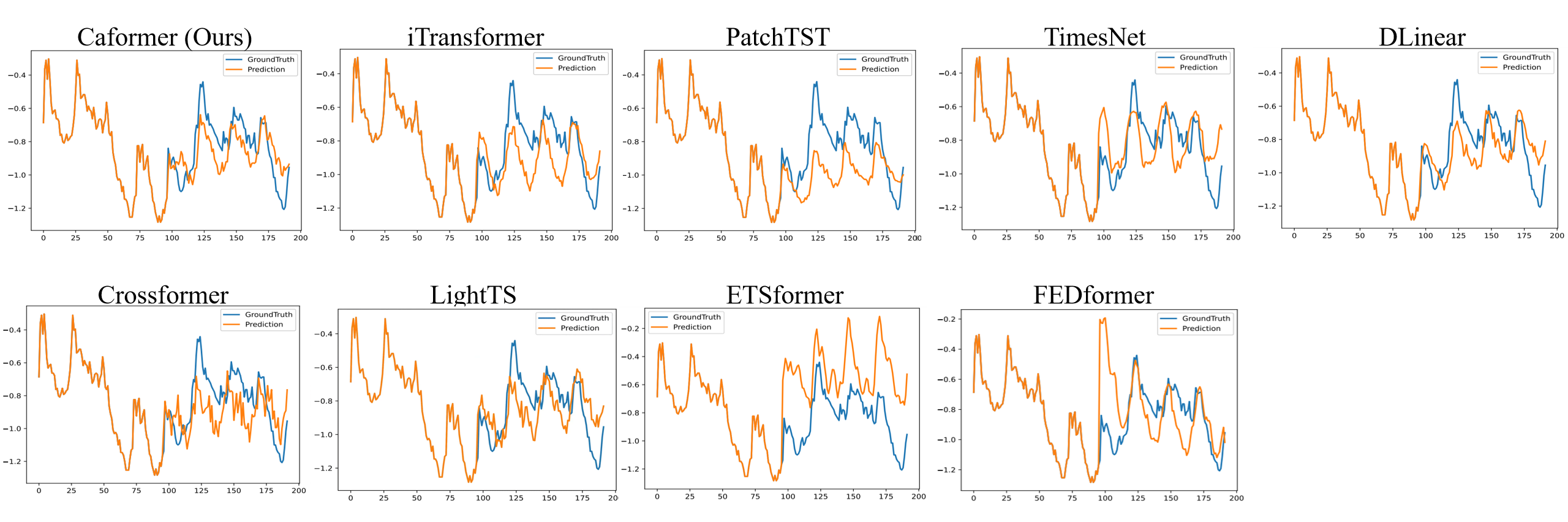}
    \caption{The visualization of long term forecasting results with ETTh1 dataset by different models under the input-96-predict-96 setting. The blue lines stand for the ground truth and the orange lines stand for the prediction.}
    \label{appendix-long}
\end{figure*}

\begin{figure*}[htbp]
    \centering
    \includegraphics[height=5cm]{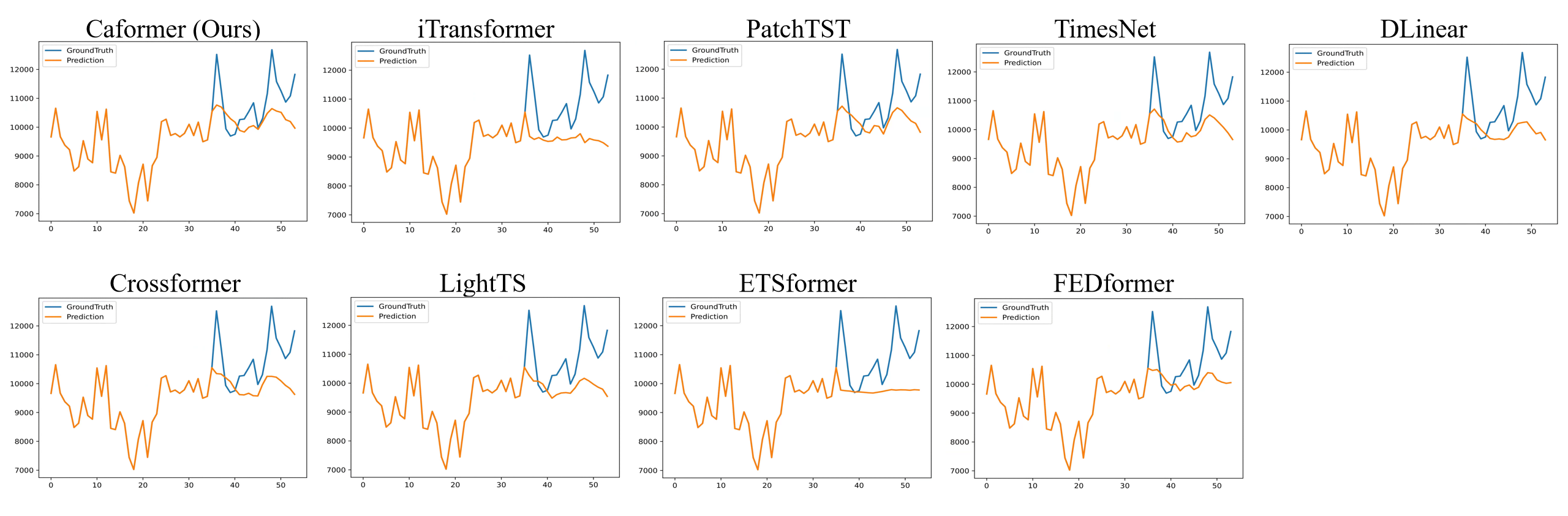}
    \caption{The visualization of imputation results with Electricity dataset by different models under the 37.5\% mask ratio setting. The blue lines stand for the ground truth and the orange lines stand for the prediction.}
    \label{appendix-imputation}
\end{figure*}

\begin{figure*}[htbp]
    \centering
    \includegraphics[height=5cm]{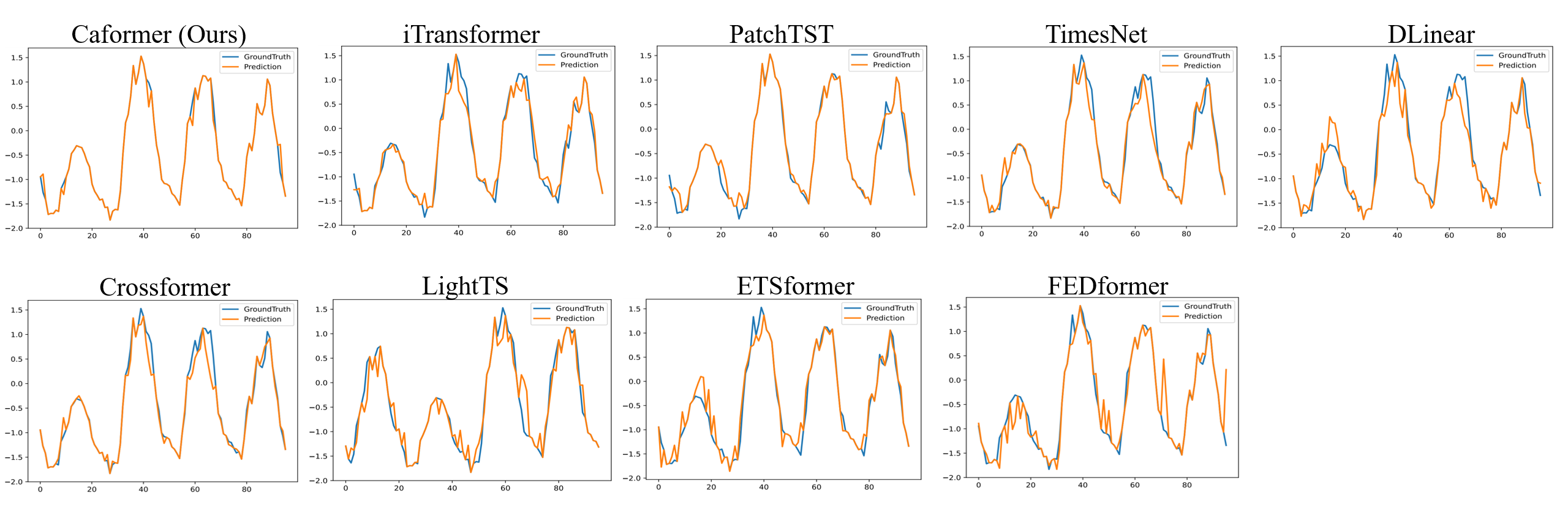}
    \caption{The visualization of short term forecasting results with the M4 dataset by different models under the monthly setting. The blue lines stand for the ground truth and the orange lines stand for the prediction.}
    \label{appendix-short}
    
\end{figure*}

\subsection{Showcases}
For a comprehensive comparison among different models, we present showcases for regression tasks, encompassing long-term forecasting as depicted in Fig. \ref{appendix-long}, short-term forecasting illustrated in Fig. \ref{appendix-short}, and imputation showcased in Fig. \ref{appendix-imputation}.
\section{Conclusion and future works}
\begin{figure}[htbp]
    \centering
    \includegraphics[height=2.6cm]{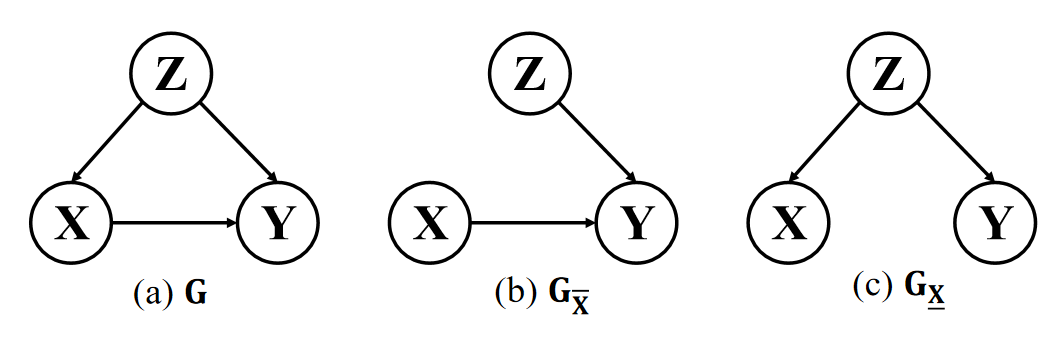}
    \caption{Illustration of different causal directed acyclic graph}
    \label{Appendixcausal}
\end{figure}

In this paper, we address the challenges of learning cross-dimension and cross-time dependencies in the presence of the environment from a causal perspective. With a comprehensive understanding of the dynamic interactions within time series, our Caformer is able to eliminate the impact of the environment while obtaining causal relationships among cross-dimension and cross-time dependencies for robust temporal representations. Experimentally, our Caformer shows great generality and performance in five mainstream analysis tasks with proper interpretability. Future investigations will delve deeper into the interactions within the time series.
\appendix[Derivation of Back-door Adjustment]
We use the back-door adjustment\cite{pearl2000models} for causal intervention. In this section, we first introduce the basic rules of do-calculus $\text{do}(\cdot)$, then we present the derivation of backdoor adjustment for the proposed causal graph in Fig. \ref{intro} based on these rules.\\
\textbf{Rules of Do-calculus.} Given an arbitrary causal directed acyclic graph $G$, there are three nodes respectively represented by $X$, $Y$, and $Z$. Particularly, $G_{\overline{X}}$ denotes the intervened causal graph with the removal of all incoming arrows leading to $X$, $G_{\underline{X}}$ denotes another intervened causal graph with the removal of all outgoing arrows from $X$. The lower cases $x$, $y$, and $z$ are used to represent the respective values of nodes: $X=x$, $Y=y$, and $Z=z$. We illustrate the details of $G$, $G_{\overline{X}}$ and $G_{\underline{X}}$ in Fig. \ref{Appendixcausal}.\\
For any interventional distribution compatible with $G$, we introduce the following three rules:\\
\textbf{Rule 1.} Insertion/deletion of observations:
\begin{equation}
    P(y|\text{do}(x), z)=P(y|\text{do}(x)),\quad \text{if}(y\upmodels z|x)_{G_{\overline{X}}},
\end{equation}
\textbf{Rule 2.} Action/observation exchange:
\begin{equation}
    P(y|\text{do}(x),\text{do}(z))=P(y|\text{do}(x),z),\quad \text{if}(y\upmodels z|x)_{G_{\overline{X}\underline{Z}}},
\end{equation}
\textbf{Rule 3.} Insertion/deletion of actions:
\begin{equation}
    P(y|\text{do}(x),\text{do}(z))=P(y|\text{do}(x)), \quad \text{if}(y\upmodels z|x)_{G_{\overline{X}\overline{Z}}}, 
\end{equation}
where $(y\upmodels z|x)_{G_{\overline{X}}}$ means that $y$ and $z$ are independent of each other given $x$ in $G$. Based on these rules, we can derive the interventional distribution $P(Y|\text{do}(T))$ for our proposed causal graph.
\begin{align}
    &\quad P(T|\text{do}(X))\\&=\sum_{c} P(T|\text{do}(X), C=c)P(C=c|\text{do}(X))\\
    &=\sum_{c}P(T|\text{do}(X), C=c)P(C=c) \\
    &=\sum_{c}P(T|X,C=c)P(C=c)
\end{align}
where (A.4) follows the law of total probability, (A.5) is obtained via Rule 3 given $cX$ in $G_{\overline{X}}$, and (A.6) can be inferred from Rule 2 which changes the intervention term into observation as $T X|c$ in $G_{\underline{X}}$.\\
By stratifying $C$ into discrete components $C=\{c_{i}\}_{i=1}^{n}$, we can finally express $P(T|\text{do}(X))$ as follows:
\begin{equation}
    P(T|\text{do}(X))=\sum_{i}^{n}P(T|X,C=c_{i})P(C=c_{i})
\end{equation}

\newpage
\bibliographystyle{IEEEtran}
\bibliography{main-new}

\end{document}